\documentclass[10pt,twocolumn,letterpaper]{article}

\usepackage{iccv}              

\usepackage{algorithm}
\usepackage[noend]{algorithmic}

\usepackage{bbding}
\usepackage{xcolor}
\usepackage{multirow}
\newcommand{\meansd}[2]{#1 $\pm$ #2}  

\usepackage{colortbl}
\definecolor{Gray}{gray}{0.90}
\newcolumntype{g}{>{\columncolor{Gray}}c}

\newcommand{\yes}{\textcolor{BrickRed}{\Checkmark}}
\newcommand{\no}{\textcolor{ForestGreen}{\XSolidBrush}}

\usepackage{amsmath,amsfonts,bm}

\def\eqref#1{equation~\ref{#1}}

\def\1{\bm{1}}

\def\vmu{{\bm{\mu}}}

\def\vc{{\bm{c}}}

\def\ve{{\bm{e}}}
\def\vf{{\bm{f}}}
\def\vg{{\bm{g}}}

\def\vl{{\bm{l}}}
\def\vm{{\bm{m}}}

\def\vt{{\bm{t}}}

\def\vv{{\bm{v}}}
\def\vw{{\bm{w}}}
\def\vx{{\bm{x}}}

\def\vz{{\bm{z}}}

\def\mC{{\bm{C}}}

\def\mE{{\bm{E}}}

\def\mG{{\bm{G}}}

\def\mL{{\bm{L}}}
\def\mM{{\bm{M}}}

\def\mT{{\bm{T}}}

\DeclareMathAlphabet{\mathsfit}{\encodingdefault}{\sfdefault}{m}{sl}
\SetMathAlphabet{\mathsfit}{bold}{\encodingdefault}{\sfdefault}{bx}{n}
\newcommand{\tens}[1]{\bm{\mathsfit{#1}}}

\def\tE{{\tens{E}}}

\def\tG{{\tens{G}}}

\def\tI{{\tens{I}}}

\def\tL{{\tens{L}}}
\def\tM{{\tens{M}}}

\def\tT{{\tens{T}}}

\def\gD{{\mathcal{D}}}

\def\gO{{\mathcal{O}}}

\def\sA{{\mathbb{A}}}
\def\sB{{\mathbb{B}}}
\def\sC{{\mathbb{C}}}

\def\sG{{\mathbb{G}}}

\def\sK{{\mathbb{K}}}

\def\sS{{\mathbb{S}}}

\newcommand{\E}{\mathbb{E}}

\newcommand{\R}{\mathbb{R}}

\newcommand{\normalize}{\mathrm{normalize}}

\DeclareMathOperator*{\argmax}{arg\,max}

\usepackage{amsthm}

\theoremstyle{plain}
\newtheorem{theorem}{Theorem}[section]

\newtheorem{lemma}[theorem]{Lemma}
\newtheorem{corollary}[theorem]{Corollary}

\theoremstyle{definition}

\newtheorem{assumption}[theorem]{Assumption}

\theoremstyle{remark}
\newtheorem{remark}[theorem]{Remark}

\newcommand{\algname}{\textit{Latte}} 

\newcommand{\dif}{\mathrm{d}} 

\newcommand{\pre}{\textnormal{pre}} 
\newcommand{\post}{\textnormal{post}} 
\newcommand{\asym}{\textnormal{asym}} 

\newcommand{\vwpre}{\vw_{\textnormal{pre}}}
\newcommand{\vwasym}{\vw_{\textnormal{asym}}}
\newcommand{\vwpost}{\vw_{\textnormal{post}}}

\newcommand{\bpre}{b_{\textnormal{pre}}}
\newcommand{\basym}{b_{\textnormal{asym}}}
\newcommand{\bpost}{b_{\textnormal{post}}}

\newcommand{\Vs}{V_{\textnormal{sphere}}}
\newcommand{\Vsc}{V_{\textnormal{sphere\_cap}}}
\newcommand{\Vol}{\textnormal{Vol}}

\newcommand{\nID}{n_\textnormal{ID}}
\newcommand{\nOOD}{n_\textnormal{OOD}}

\renewcommand{\paragraph}[1]{\vspace{5pt} \noindent\textbf{#1.}\ }

\definecolor{iccvblue}{rgb}{0.21,0.49,0.74}
\usepackage[pagebackref,breaklinks,colorlinks,allcolors=iccvblue]{hyperref}

\def\paperID{277} 
\def\confName{ICCV}
\def\confYear{2025}

\title{{\algname}: Collaborative Test-Time Adaptation of Vision-Language Models in Federated Learning}

\author{
Wenxuan Bao \quad Ruxi Deng \quad  Ruizhong Qiu \quad  Tianxin Wei \quad  Hanghang Tong \quad  Jingrui He \\
University of Illinois Urbana-Champaign \\
\tt\small \{wbao4,ruxid2,rq5,twei10,htong,jingrui\}@illinois.edu
}

\begin{document}
\maketitle

\addtocontents{toc}{\protect\setcounter{tocdepth}{0}} 

 \begin{abstract}
    Test-time adaptation with pre-trained vision-language models has gained increasing attention for addressing distribution shifts during testing. Among these approaches, memory-based algorithms stand out due to their training-free nature and ability to leverage historical test data. However, existing test-time adaptation methods are typically designed for a single domain with abundant data. In decentralized settings such as federated learning, applying these methods individually to each client suffers from limited test data, while directly sharing a single global memory via the server prevents proper personalization to each client's unique distribution.
    To address this, we propose {\algname}, a novel framework where each client maintains a local memory to store embeddings from its own historical test data and an external memory to store class prototypes from other relevant clients. During communication, each client retrieves prototypes from similar clients under the server’s coordination to expand its memory. For local adaptation, {\algname} utilizes both embedding similarity and uncertainty to enhance model performance.
    Our theoretical analysis shows that {\algname} effectively leverages in-distribution clients while remaining robust to out-of-distribution clients. Extensive experiments on domain adaptation and corruption benchmarks validate that {\algname} achieves superior performance in decentralized settings, while introducing only negligible communication and computation costs. 
    Our code is available at \url{https://github.com/baowenxuan/Latte}. 
\end{abstract}
\section{Introduction}

Pre-trained vision-language models (VLMs), such as CLIP \cite{clip} and ALIGN \cite{align}, have demonstrated impressive zero-shot image classification capabilities by being trained on massive datasets of image-text pairs. However, when these models are deployed in specific downstream domains, the presence of domain shifts can lead to a misalignment between vision and text embeddings. This misalignment may affect the model's classification performance, as the learned embeddings may no longer fully capture the nuanced relationships required in the target domain \cite{coop,cocoop,tip-adaptor}. 

Recently, a number of approaches have applied test-time adaptation (TTA) to mitigate domain shift challenges. TTA enables models to adapt directly to a stream of unlabeled target data samples, without accessing source data, using techniques such as prompt tuning \cite{tpt,difftpt,histpt}, prompt weighting \cite{zpe}, entropy minimization \cite{tps,dpe}, distribution alignment \cite{promptalign}, etc. Among these, memory-based algorithms \cite{tip-adaptor,tda,dmn,dpe} have emerged as an important category. These methods store high-confidence test image embeddings and pseudo-labels in a memory, which is then used to adjust the classifier based on the similarity between the new image's embedding and stored ones. Notably, memory-based algorithms are training-free and require no backpropagation, making them highly efficient and particularly suitable for scenarios with limited computational resources. Moreover, by leveraging the similarity among test samples, the performance of these methods often improves as the amount of test data increases \cite{dpe}, making them well-suited for adapting to target domains with abundant data. 

Despite their promising performance on benchmark datasets, existing memory-based TTA approaches primarily focus on adaptation to a single domain with abundant data. However, many real-world applications require adapting VLMs to a series of related but distinct domains, where each domain may have very limited data. One prominent example is federated learning (FL) \cite{fl_survey_advance,fedavg}, where multiple clients perform the same task but each client has access to only a small amount of data drawn from its unique distribution \cite{atp,fedthe}. As shown in Figure \ref{fig:num_clients}, when data becomes more distributed, applying existing memory-based TTA algorithms independently to each client overlooks the relationships among clients and results in performance degradation due to data scarcity. On the other hand, directly sharing a single, unified memory across clients can prevent the VLM from being effectively personalized to individual clients, as each client’s distribution may be different. 

To address these challenges, we propose {\algname}, a colLAborative Test-TimE adaptation algorithm designed for VLMs in FL settings. In {\algname}, each client maintains both a local memory and an external memory. The local memory stores embeddings of high-confidence images from the client’s own distribution, while the external memory contains prototypes shared by relevant clients. This design effectively balances cross-client knowledge sharing with personalization for individual clients. 
During communication, each client aggregates its local memory into class prototypes, uploads them to the server, and uses them as keys to retrieve prototypes from other relevant clients with in-distribution (ID) samples, which are then stored in the external memory. 
For adaptation, to improve robustness against out-of-distribution (OOD) samples and misclassified samples, {\algname} leverages both embedding similarity and entropy to weight embeddings in the memory, thereby enhancing model performance. 
To ensure practical usability, our approach decouples local test-time adaptation from communication, allowing clients to perform offline inference while significantly improving communication efficiency. 
We summarize our contributions as follows:
\begin{itemize}
    \item We propose a novel TTA algorithm {\algname}, a collaborative test-time adaptation algorithm for VLM in FL. {\algname} includes local and external memories to balance cross-client knowledge sharing with personalization. while the adaptation is robust to OOD and misclassified samples. 
    \item We conduct a comprehensive theoretical analysis of {\algname}, covering both memory construction and adaptation, and prove that {\algname} effectively leverages ID samples while remaining robust to OOD samples.
    \item We evaluate {\algname} on both domain adaptation and corruption benchmarks, where {\algname} consistently outperforms existing baselines while incurring negligible additional computational and communication costs. 
\end{itemize}

\section{Related Works}

In this section, we summarize related works of test-time adaptation in both centralized and federated learning, with a focus on vision-language models. Additionally, we provide a broader discussion of related works in Appendix \ref{appendix:discussion:related_works}. 

\paragraph{Test-time adaptation}
Test-time adaptation (TTA) aims to adapt a pre-trained model from the source domain to an unlabeled target domain without requiring access to the source data \cite{tta_survey,tta_survey_2}. This is particularly suitable for pre-trained vision-language models (VLMs) such as CLIP \cite{clip}, where only the model weights are available. Early TTA approaches for VLMs primarily focused on prompt tuning: They generate multiple augmented views and learn instance-specific prompts for each image via entropy minimization \cite{tpt,difftpt,histpt}, distribution alignment \cite{promptalign}. While these approaches have demonstrated effectiveness across various datasets, the use of data augmentation and backpropagation introduces significant computational overhead. Training-free TTA methods \cite{augmix,tps,zero} utilizes data augmentations and confidence selection to improve the test-time robustness. However, they still requires model inference on multiple augmented images, which introduce significant computation cost. 
Recently, inspired by Tip-Adapter \cite{tip-adaptor}, memory-based TTA methods \cite{tda,dmn,dpe,boostadapter} store the embeddings and pseudo-labels of high-confidence test samples in a memory. When making predictions for new test images, they use embedding similarities to refine model prediction and enhance the performance. 

\paragraph{Federated test-time adaptation}
Federated test-time adaptation (FTTA) requires personalizing a global model to a series of FL clients without accessing raw data from other clients. Some FTTA algorithms address the challenge of distribution shifts between training and testing phases for individual clients. For example, FedTHE \cite{fedthe} dynamically aggregates the global and local models, while FedICON \cite{fedicon} leverages inter-client heterogeneity to enhance model robustness against distribution shifts. Another line of works focuses on generalization to new clients. ATP \cite{atp} achieves adaptive personalization by learning module-wise adaptation rates, whereas FedCal \cite{fedcal} targets label shifts by estimating the testing label distribution to calibrate the FL model. However, most of these methods are designed for models trained within FL settings and are not applicable to pre-trained VLMs. Additionally, during testing, no information is exchanged between clients. The most similar setting to our work is TSA \cite{tsa}, which aggregates the locally adapted models and uses the aggregated model for testing, enabling collaborative test-time adaptation. However, it focuses on a special temporal-spatial shift scenario, which may not be directly applicable in our setting. 

\section{Method}

\begin{figure*}
    \centering
    \includegraphics[width=0.98\linewidth]{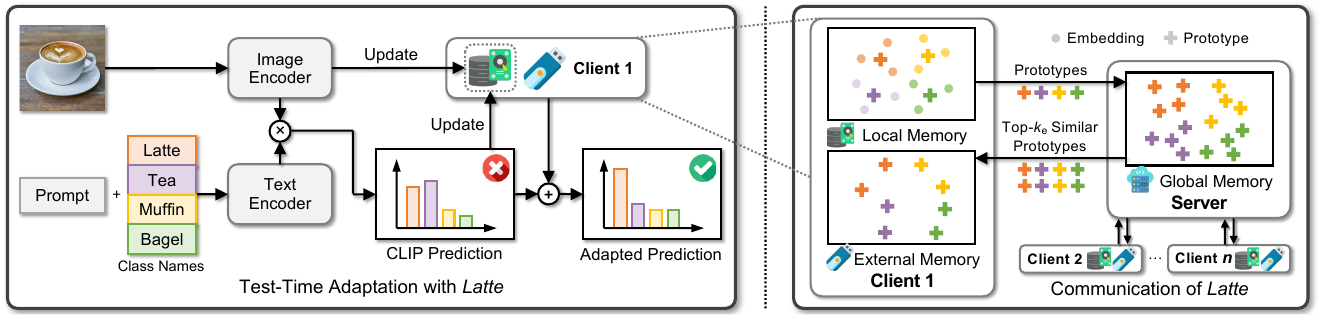}
    \caption{Overview of our proposed framework {\algname}. (Left) Each client performs test-time adaptation using local and external memory. (Right) Clients communicate with the server to update external memory}
    \label{fig:framework}
\end{figure*}

Following \cite{atp,tsa,cola}, we consider an FL system with $n$ clients, indexed by $i = 1, \cdots, n$, where each client has a data distribution $\gD_i$ that generates a stream of test images. All clients share the same feature space and label space, while the distributions $\gD_i, \gD_j$ for two clients may be identical, similar, or significantly different. The objective of {\algname} is to maximize the average testing accuracy across all clients, while minimizing additional computational and communication costs. For each client $i$, when encountering a new testing image, {\algname} follows four steps: 
\begin{enumerate}
    \item Encode an input image into image embedding $\vf$ and get initial prediction. (Subsection \ref{subsec:prelim})
    \item Update the local memory $\tL^i$ with $\vf$. (Subsection \ref{subsec:local_memory})
    \item Use local memory $\tL^i$ and external memory $\tE^i$ to get adapted prediction. (Subsection \ref{subsec:adapt})
    \item (Optional) Communicate with the server and update the external memory $\tE^i$. (Subsection \ref{subsec:external_memory})
\end{enumerate}
Figure \ref{fig:framework} provides an overview of the algorithm, while the pseudo-code is presented in Appendix \ref{appendix:method:pseudo_code}. For readability, we additionally summarize the notations in Appendix \ref{appendix:method:notation}. 


\subsection{Preliminaries}
\label{subsec:prelim}

\textbf{CLIP} \cite{clip} is a VLM consisting of an image encoder and a text encoder, which aligns images with their corresponding textual descriptions. Pretraining on a large-scale image-text dataset enables CLIP to perform zero-shot prediction. Specifically, CLIP's learned text encoder synthesizes a zero-shot linear classifier by embedding the names or descriptions of the target dataset’s classes (e.g., ``\texttt{a photo of a \{class\}}'') into (normalized) text embeddings, represented as $\mT = [\vt_1, \cdots, \vt_c]^\top \in \R^{c \times d}$, where $c$ is the number of classes and $d$ is the dimension of embeddings. 
Given a new test image, CLIP uses the image encoder to encode it into (normalized) image embedding $\vf$. The predicted logits are computed as $\vz_{\pre} = 100 \cdot \vf^\top \mT^\top$, and the final prediction is given by $\argmax \vz_{\pre}$. Here $\vf$ denotes the image embedding, and $\mT$ is the matrix of text embeddings, both L2-normalized. 

\textbf{Memory-based TTA} \cite{tda,dmn,dpe} is a kind of online TTA approach that enhances CLIP's performance on downstream tasks. It maintains a memory (cache) of past test image embeddings and their pseudo-labels (predictions), and uses them to adjust the prediction of new images based on embedding similarities. This method benefits from previously seen test images, making it particularly suitable for scenarios with large sample sizes. Additionally, since these algorithms often do not require backpropagation, they offer high computational efficiency.


\subsection{Construct Local Memory}
\label{subsec:local_memory}

To construct local memory for each client, we use a standard method based on priority queue \cite{tda,dmn,dpe,adaprompt}. For client $i$, we construct a class-split memory $\tL^{i} \in \R^{c \times k_l \times d}$, where $k_l$ is the memory size for each class. The memory consists of $c$ independent priority queues $\{\mL_1^i, \cdots, \mL_c^i\}$, one for each class. Each $\mL_y^i = \{\vl_{y, 1}^i, \cdots, \vl_{y, k_l}^i\}$ stores test image embeddings associated with the corresponding pseudo-class $y$, sorted from low entropy to high entropy. The queues are initially empty. When a new test image arrives, we first use CLIP to obtain its image embeddings $\vf$, pseudo-label $\hat{y}$, and the corresponding entropy $H(\vf) = - \sum_{y = 1}^c \hat{p}_{y} \log \hat{p}_y$, where $\hat{p}_y = \frac{\exp(100 \cdot \vf^\top \vt_y)}{\sum_{y'=1}^c \exp(100 \cdot \vf^\top \vt_{y'})}$. 
We then update $\mL_{\hat{y}}^i$, the queue corresponding to $\hat{y}$: 
\begin{itemize}
    \item If $\mL_{\hat{y}}^i$ is not full, the new test image embedding is directly inserted.
    \item If $\mL_{\hat{y}}^i$ is full, the new embedding replaces the entry with the highest entropy in the queue.
\end{itemize}
This process can be efficiently implemented using heaps. Since entropy estimates the uncertainty of a prediction, a lower entropy value suggests a more reliable pseudo-label. Intuitively, as more samples are observed, the memory stores increasingly reliable samples, which in turn enhances the prediction of new test samples.


\subsection{Communication among Clients}
\label{subsec:external_memory}

The effectiveness of memory-based algorithms relies on the local memory containing a sufficient number of high-quality embeddings. However, when testing samples are distributed across many different clients, with each client having only a small number of testing samples, the local memory may lack adequate high-quality embeddings to support enhanced prediction. To address this, it is necessary to exchange memory information among clients, thereby enhancing the overall utility of the memory. 

A straightforward approach is to maintain a shared global memory on the central server, with all clients staying synchronized with the server’s global memory. This approach simulates a scenario where data is aggregated into a centralized repository. However, FL clients often exhibit heterogeneous data distributions. Sharing a global memory would make each client's memory identical, forcing all client models to behave uniformly. This limits the degree of personalization and hinders the ability to fully adapt to the unique distribution of each client. Instead, in our proposed {\algname}, while the server maintains a global memory that captures diverse client information, each client maintains its own external memory for test-time adaptation. 

\paragraph{Global memory}
In {\algname}, the server maintains a class-split global memory $\tG = \{\mG_1, \cdots, \mG_c\} \in \R^{c \times n \times d}$, where $n$ represents the number of clients. For each class $y$, the corresponding segment of global memory $\mG_y = [\vg_{y, 1}, \cdots, \vg_{y, n}]^\top \in \R^{n \times d}$ stores one prototype from each client, serving as a representative summary of its local memory. It ensures the global memory stores information from every client. For each client $i$, after updating its local memory and completing the prediction, if it is connected to the server and ready to communicate, it can compute its prototype based on the latest local memory and upload it to the server as follows: 
\begin{align}
    \vg_{y, i} \xleftarrow{\text{upload}} \normalize\left( \sum_{\kappa=1}^{k_l} \exp(- \gamma \cdot H(\vl^i_{y, \kappa})) \cdot \vl^i_{y, \kappa}\right),
    \label{eq:global_memory}
\end{align}
where $\normalize(\vx) = \frac{\vx}{\| \vx \|_2}$ is the $L^2$ normalization, and $\gamma$ is a sharpness hyperparameter. The prototype is the weighted average of embeddings in the local memory for a given class $y$. We assign higher weights to embeddings with higher certainty (i.e., lower entropy) to mitigate the influence of misclassified samples in memory and enhance the reliability of the prototype. 

\paragraph{External memory}
After constructing the global memory, directly sharing it with each client may be ineffective due to distribution differences across clients. For a given client $i$, only a small subset of clients' prototypes may be in-distribution (ID), i.e., sharing the same or highly similar distribution, while the majority may be out-of-distribution (OOD). These OOD prototypes provide little to no benefit for classification and, if overly abundant, may even degrade performance. Moreover, transmitting these irrelevant prototypes introduces unnecessary communication costs. To address this, for each client $i$, we treat the uploaded prototype $\vg_{y, i}$ as a query vector to retrieve the top-$k_e$ most similar prototypes from the global memory. The retrieved prototypes are then stacked and returned to the client as the external memory $\tE^i = \{\mE_1^i, \cdots, \mE_c^i\} \in \R^{c \times k_e \times d}$, reducing the transmission of irrelevant prototypes. Formally, this process can be expressed by
\begin{align}
    \mE_y^i \xleftarrow{\text{download}} \{\vg_{y, j} \in \mG_y: j \neq i, \vg_{y, j}^\top \vg_{y, i} \geq \tau(k_e)\}, 
\end{align}
where $\tau(k_e)$ is a threshold selecting $k_e$ vectors with the highest cosine similarities to $\vg_{y, i}$, excluding $\vg_{y, i}$ itself. It reduces the download communication cost from $cnd$ to $c k_e d$ parameters. Note that this step serves only as a coarse filtering of prototypes. During inference, an adaptive aggregation will be performed based on the new test image, which we will introduce in the next subsection. 

Unlike other collaborative TTA algorithms such as TSA \cite{tsa}, our communication process depends only on each client’s local memory $\tL^i$ and is independent of the current test sample $\vf$. This decouples communication from local TTA, allowing communication frequency to be adjusted based on network conditions rather than for every new sample, significantly reducing communication rounds. Moreover, clients can perform offline inference using their local and previously updated external memory, making our approach more practical and suitable for general FL systems.


\subsection{Adaptation with Local and External Memories}
\label{subsec:adapt}

Finally, we introduce how each client in {\algname} makes test-time adaptation, based on its local and external memories. We first merge the local memory $\tL^i$ and external memory $\tE^i$ into a merged memory $\tM^i$. For each $y$, 
\begin{align}
    \mM^i_y = \left \{\vm \in \mL^i_y \cup \mE^i_y : H(\vm) \leq \tau(k_l) \right\}, 
\end{align}
where $\tau(k_l)$ is a threshold ensuring that only the $k_l$ vectors with the lowest entropy are selected. If the prototypes in the external memory have higher certainty, they will replace the least certain embeddings in the local memory, thereby improving the overall memory quality. However, it also introduces the potential risk of OOD prototypes, especially when $k_e$ is large. Simply averaging the prototypes in memory may be overly sensitive to OOD prototypes. Additionally, in the early stages of TTA, the memory may contain both correctly and incorrectly classified embeddings/prototypes, where the misclassified samples can have a significantly negative impact on performance. Therefore, when computing the similarity from a new test sample $\vf$ to the merged memory $\mM_y^i$ for each class $y$, we need to consider both embedding similarity and uncertainty. Inspired by cross-attention \cite{dmn}, we compute the memory logits as follows: 
\begin{align}
    w^i_{y, \kappa} &= \exp \left( \beta \cdot \vf^\top \vm^i_{y, \kappa}\right) \cdot \exp \left( - \gamma \cdot H \left(\vm^i_{y, \kappa} \right) \right), \\
    \vc^i_y &= \normalize \left( \sum_{\kappa=1}^k w^i_{y, \kappa}\cdot \vm^i_{y, \kappa} \right), \\
    \vz_{\text{mem}} &= 100 \cdot \vf^\top \mC^i,  \text{ where } \mC^i = \left[\vc^i_1, \cdots, \vc^i_c\right]. 
\end{align}
Here, the aggregation weight $w^i_{y, \kappa}$ is controlled by both embedding similarity $\vf^\top \vm^i_{y, \kappa}$ and uncertainty $H(\vm^i_{y, \kappa})$, with $\beta$ and $\gamma$ as hyperparameters that regulate the sharpness of these effects. By assigning higher weights to samples with high similarity and low uncertainty, $\vc_y^i$ becomes more robust to both OOD prototypes from other clients and uncertain embeddings. Intuitively, for a given class $y$, if there exist one or a few samples that are both highly similar to the new test image embedding and highly certain, we assign the new sample to this class while allowing the memory to retain low-similarity or high-uncertainty samples. This design enables our memory adapter to benefit from additional in-distribution (ID) client data while remaining robust to OOD clients. We provide a theoretical analysis of this property in Theorem \ref{thm:multi_domain} in the next section.

Finally, the final prediction is obtained by aggregating the CLIP prediction and the memory prediction: 
\begin{align}
    \hat{y}_{\post} = \argmax\vz_{\post}, \text{ where }\vz_{\post} = \vz_{\pre} + \alpha \cdot \vz_{\text{mem}}, 
\end{align}
where $\alpha$ is a positive hyperparameter. 

\section{Theoretical Analysis}

In this section, we analyze the generalization performance of {\algname}. In this direction, while prior theoretical analysis \cite{adanpc,boostadapter} extended $k$-NN generalization analysis from the supervised setting to test-time adaptation, they relied on assumptions of strong density and noise-free memory, which may not hold in TTA at the same time. Moreover, they evaluated memory classifiers in isolation, without comparing them to the baseline zero-shot classifier, making it difficult to assess the advantages of TTA. In contrast, we propose a framework that accounts for memory construction and enables direct comparison with zero-shot classifier. Using this framework, we further show that {\algname} benefits from ID clients while remaining robust to highly OOD clients. \textit{The formal assumptions, theorems and complete proofs can be found in Appendix \ref{appendix:proof}. }

\paragraph{Setting and assumptions}
For simplicity, we consider a binary classification problem, where $y \in \{0, 1\}$. The image embeddings for each class $y$ are assumed to be drawn uniformly from a $d$-dimensional unit sphere with center $\vmu_y = (2y -1) \cdot \vmu$. The initial linear zero-shot classifier is given by $\hat{y} = 1 \{ \vwpre^\top \vf + \bpre > 0 \}$, with an associated error rate $\epsilon_\pre$. Additionally, we assume that $\vwpre^\top \vmu > 0$ and $\epsilon_\pre < 1/2$. This assumption is minimal and serves to rule out degenerate cases where the initial classifier completely flips the class assignments or collapses into predicting a single class for all samples. 

\begin{lemma}[Single client asymptotic generalization] \label{lem:asymptotic}
    Applying {\algname} to a single client. Suppose {\algname} has shot capacity $k$ and hyperparameters $\alpha,\beta \to + \infty$. When the number of seen testing samples $N \to + \infty$, the expected error rate on unseen testing samples converges to $\epsilon_{\asym} \leq \epsilon_{\pre}$. 
\end{lemma}


\begin{theorem}[Single client generalization] \label{thm:single_domain}
    Under the same assumptions of Lemma \ref{lem:asymptotic}, with probability at least $1-\delta$, for sufficiently large $N$, the expected error rate on unseen testing samples satisfies 
    \begin{align}
        \epsilon_\post \leq \epsilon_{\asym} + \gO \left( \left( \frac{k+\log\frac1\delta}{N} \right)^{\frac{1}{d+1}} \right), 
    \end{align}
    where $k$ is the merged memory size. Moreover, when $\epsilon_{\asym} = 0$, a faster convergence w.r.t. $N$ is guaranteed:
    \begin{align}
        \epsilon_\post \leq \gO \left( \left( \frac{k+\log\frac1\delta}{N} \right)^{\frac{1}{2}} \right). 
    \end{align}
\end{theorem}


Intuitively, Lemma \ref{lem:asymptotic} and Theorem \ref{thm:single_domain} indicate that as the number of ID samples increases, {\algname}'s expected error rate progressively decreases, and eventually reaches a level superior to that of the zero-shot classifier. This implies that, as a memory-based method, {\algname}'s performance continually improves with an increasing number of test images. This highlights the need for memory-based methods to have access to a larger number of test images for optimal performance. 

Next, we consider applying {\algname} to multiple clients, and analyze the performance on one client $i$. Suppose for each client, there are $\nID$ clients with the same distribution, and $\nOOD$ clients with different distribution. Assume OOD clients follow the same assumption as ID clients, except with different sphere centers $\vmu_y' \neq \vmu_y$. We assume $\forall y, y' \in \{0, 1\}$, $\| \vmu_{y'}' - \vmu_y \|_2 > 4$, which indicates that different distributions are distinctly different. 

\begin{theorem}[Multiple clients generalization] \label{thm:multi_domain}
    Under the same assumptions of Theorem \ref{thm:single_domain}, suppose the number of seen testing samples for each client is $N$. With probability at least $1-\delta$, for sufficiently large $N$, the expected error rate on unseen testing samples satisfies 
    \begin{align}
        \epsilon_\post \leq \epsilon_{\asym} + \gO \left( \left( \frac{k+\log\frac1\delta}{\nID \cdot N} \right)^{\frac{1}{d+1}} \right), 
    \end{align}
    where $k$ is the merged memory size. Moreover, when $\epsilon_{\asym} = 0$, a faster convergence w.r.t. $N$ is guaranteed:
    \begin{align}
        \epsilon_\post \leq \gO \left( \left( \frac{k+\log\frac1\delta}{\nID \cdot N} \right)^{\frac{1}{2}} \right). 
    \end{align}
\end{theorem}

We observe that the error bound decreases as $\nID$ increases but remains unaffected by the growth of $\nOOD$. This confirms that {\algname} benefits from ID clients while remaining robust to highly OOD clients.

\section{Experiments}

\begin{table*}
    \centering
    \scriptsize
        \begin{tabular}{lcccccgcccccg}
            \toprule
            \multirow{2.5}{*}{Method} & \multicolumn{6}{c}{VLCS} & \multicolumn{6}{c}{TerraIncognita}  \\
            \cmidrule(lr){2-7} \cmidrule(lr){8-13} 
            & C & L & S & V & Total & \cellcolor{White}Gain & L100 & L38 & L43 & L46 & Total & \cellcolor{White}Gain \\
            \midrule
            CLIP (ViT-B/16) \cite{clip} 
                & 99.86 & 70.11 & 76.66 & 85.34 & 80.83 & -
                & 41.96 & 28.30 & 35.82 & 26.82 & 31.84 & - \\
            VTE \cite{vte}            
                & 99.86 & 70.29 & 78.28 & 86.55 & 81.75 & +0.92
                & 46.81 & 35.25 & 47.86 & 31.17 & 38.56 & +6.72\\
            TPT \cite{tpt} 
                & 99.93 & 66.64 & 79.37 & 86.52 & 81.18 & +0.35
                & 57.18 & 28.48 & 36.97 & 28.49 & 35.51 & +3.67\\
            Zero \cite{zero}
                & 99.86 & 70.29 & 78.85 & 86.73 & 81.98 & +1.15
                & 45.56 & 35.20 & 47.66 & 31.48 & 38.34 & +6.50\\
            TDA (local) \cite{tda} 
                & 99.94 & 71.05 & 77.82 & 85.39 & 81.44 & +0.61
                & 40.46 & 32.72 & 36.23 & 30.41 & 34.24 & +2.40\\
            TDA (global) \cite{tda} 
                & 99.94 & 67.53 & 76.95 & 85.34 & 80.29 & -0.54
                & 41.78 & 39.58 & 35.37 & 26.87 & 36.19 & +4.35\\
            DMN-ZS (local) \cite{dmn}
                & 99.90 & 70.50 & 77.58 & 85.05 & 81.12 & +0.26
                & 37.79 & 31.37 & 40.70 & 29.34 & 33.65 & +1.81\\
            DMN-ZS (global) \cite{dmn}
                & 99.93 & 68.03 & 77.75 & 85.00 & 80.55 & -0.28
                & 38.33 & 44.39 & 36.75 & 26.91 & 37.64 & +5.80\\
            {\algname}
                & 99.96 & 73.00 & 80.18 & 85.13 & 82.57 & \textbf{+1.74}
                & 39.54 & 51.54 & 33.43 & 30.12 & 40.95 & \textbf{+9.11}\\
            \midrule
            \midrule

            CLIP (RN50) \cite{clip} 
                & 99.36 & 68.60 & 73.52 & 84.92 & 79.30 & -
                & 14.70 & 20.04 & 34.43 & 27.82 & 23.24 & - \\
            VTE \cite{vte}            
                & 99.58 & 66.75 & 74.80 & 84.69 & 79.19 & -0.11
                & 29.74 & 36.62 & 32.01 & 28.29 & 32.46 & +9.22\\
            TPT \cite{tpt} 
                & 99.43 & 65.25 & 77.39 & 86.43 & 80.14 & +0.84
                & 15.21 & 38.30 & 31.79 & 24.50 & 29.29 & +6.05\\
            Zero \cite{zero}
                & 99.65 & 66.42 & 74.59 & 84.69 & 79.05 & -0.25
                & 26.11 & 33.77 & 31.07 & 26.79 & 30.10 & +6.86\\
            TDA (local) \cite{tda} 
                & 99.56 & 69.32 & 74.59 & 84.77 & 79.78 & +0.48
                & 29.68 & 40.39 & 38.80 & 35.14 & 36.73 & +13.49
                \\
            TDA (global) \cite{tda} 
                & 99.59 & 67.45 & 74.16 & 84.56 & 79.12 & -0.18
                & 29.28 & 43.16 & 39.58 & 34.80 & 37.78 & +14.54
                \\

            DMN-ZS (local) \cite{dmn}
                & 99.48 & 69.05 & 73.92 & 84.66 & 79.47 & +0.17
                & 25.81 & 32.52 & 37.45 & 32.07 & 31.89 & +8.65
                \\
            DMN-ZS (global) \cite{dmn}
                & 99.56 & 65.89 & 74.77 & 83.96 & 78.73 & -0.57
                & 25.68 & 44.35 & 36.71 & 29.33 & 35.73 & +12.49
                \\
            {\algname} 
                & 99.66 & 72.07 & 77.19 & 83.12 & 80.75 & \textbf{+1.45}
                & 47.51 & 45.18 & 37.38 & 33.40 & 41.47 & \textbf{+18.23}
                \\
            
            \bottomrule 
        \end{tabular}
    \caption{Accuracy (\%, mean over five random seeds) on domain adaptation benchmarks.}
    \label{tab:acc:domain}
\end{table*}

In this section, we use experiments to answer the following questions: 
\begin{itemize}
    \item \textbf{RQ1}: Is {\algname} effective with decentralized and heterogeneous clients? 
    \item \textbf{RQ2}: Is {\algname} efficient in both communication and computation? 
    \item \textbf{RQ3}: How clients in {\algname} share memories, and how these external memories improve memory qualities? 
\end{itemize}

\paragraph{Datasets}
Following previous works in centralized and federated TTA \cite{adanpc,tent,atp}, we test our proposed algorithm on domain adaptation benchmarks (VLCS \cite{vlcs}, TerraIncognita \cite{terra-incognita}) and corruption benchmarks (CIFAR-10-C, CIFAR-100-C \cite{cifar,corruption}).\footnote{Since our focus is on adaptation to multiple target clients that share the same task but have different distributions, OOD benchmarks and cross-domain benchmarks used in \cite{clip,tpt} are not suitable for our evaluation.} 
We adopt a data partitioning strategy similar to \cite{atp}: For domain adaptation benchmarks, each domain's dataset is evenly distributed to $m=10$ clients, resulting in $n=40$ clients for both datasets. For corruption benchmarks, we first partition the original CIFAR-10/100 into $19 \cdot m$ clients. Then, for every $m$ clients, we apply a single corruption type with severity level 5, ensuring that different augmentations of the same image do not appear within the same dataset. We use $m=10$ for CIFAR-10 and $m=3$ for CIFAR-100. We summarized the statistics of these datasets in Table \ref{tab:dataset_stats}. 

\paragraph{Baselines}
We mainly compare {\algname} to other memory-based TTA methods designed for VLM, i.e., TDA \cite{tda} and DMN-ZS \cite{dmn}. For each of them, we develop a ``local'' and a ``global'' version. The local version run the TTA method on each client independently, without sharing any information among them; while the global version simply share the memory for all clients. Besides memory-based TTA methods, we also compare to episodic TTA algorithms: VTE \cite{vte}, TPT \cite{tpt}, and Zero \cite{zero}. These methods conduct TTA on each image independently, without using any information from other images. Instead, they use AugMix \cite{augmix} augmentation and confidence selection to improve the test-time robustness to distribution shifts. 

\paragraph{Settings}
We use visual encoders of ResNet \cite{resnet} (RN50) and Vision Transformer \cite{vit} (ViT-B/16) pretrained by CLIP. Following previous works \cite{tip-adaptor,tpt,tda,dmn}, for all baselines except TPT who tunes the prompt, we ensemble the text embeddings of seven different templates given by \cite{tip-adaptor} as the initial classifier. We do not apply AugMix to memory-based TTA algorithms, including {\algname}.  We use the same hyperparameter selection strategy in \cite{dmn} for all baselines, and further illustrate {\algname}'s robustness to hyperparameter selection. More details of experiments setup is provided in Appendix \ref{appendix:exp:setup}. 


\begin{table*}
    \centering
    \small
    \resizebox{1.0\linewidth}{!}{
    \setlength{\tabcolsep}{1.0mm}{
        \begin{tabular}{lccccccccccccccccccccg}
            \toprule
            & \multicolumn{20}{c}{CIFAR-10-C} \\
            \cmidrule(lr){2-22}
            Method & \multicolumn{4}{c}{Noise} & \multicolumn{5}{c}{Blur} & \multicolumn{5}{c}{Weather} & \multicolumn{5}{c}{Digital} & \multirow{2.5}{*}{Total} & \multirow{2.5}{*}{\cellcolor{White}Gain} \\
            \cmidrule(lr){2-5} \cmidrule(lr){6-10} \cmidrule(lr){11-15} \cmidrule(lr){16-20}
            & Gauss. & Shot & Impulse & Speckle 
            & Defocus & Glass & Motion & Zoom & Gauss. 
            & Snow & Frost & Fog & Bright & Spatter
            & Contrast & Elastic & Pixel & JPEG & Saturate \\
            \midrule
            CLIP (ViT-B/16) \cite{clip} 
                & 40.89 & 45.36 & 58.14 & 48.66 & 73.20 & 44.20 & 70.21 & 75.33 & 71.30 & 76.34 & 79.88 & 72.12 & 85.93 & 84.38 & 65.56 & 55.93 & 51.05 & 61.59 & 85.93 & 65.58 & - \\
            VTE \cite{vte}            
                & 44.60 & 49.62 & 63.21 & 51.67 & 71.21 & 43.81 & 70.98 & 73.73 & 68.86 & 76.69 & 79.59 & 73.95 & 84.90 & 82.05 & 82.89 & 59.02 & 60.84 & 61.92 & 85.62 & 67.64 & +2.06\\
            TPT \cite{tpt} 
                & 40.17 & 45.53 & 59.87 & 48.17 & 72.68 & 43.43 & 70.88 & 75.19 & 70.66 & 76.14 & 78.71 & 73.22 & 84.46 & 82.77 & 71.00 & 57.89 & 53.02 & 61.47 & 85.17 & 65.81 & +0.23\\
            Zero \cite{zero}
                & 44.96 & 50.02 & 63.46 & 51.90 & 71.58 & 44.68 & 71.05 & 73.86 & 69.08 & 76.98 & 79.74 & 74.27 & 85.18 & 82.46 & 82.75 & 59.85 & 60.98 & 62.98 & 85.36 & 67.95 & +2.37\\
            TDA (local) \cite{tda} 
                & 40.72 & 45.85 & 59.85 & 48.83 & 74.23 & 45.54 & 71.38 & 76.61 & 72.28 & 77.19 & 80.55 & 73.11 & 86.39 & 84.94 & 67.68 & 57.97 & 52.84 & 62.52 & 86.43 & 66.58 & +1.00\\
            TDA (global) \cite{tda} 
                & 39.46 & 44.81 & 56.56 & 47.93 & 73.05 & 44.80 & 70.67 & 75.79 & 70.94 & 76.90 & 80.53 & 72.65 & 86.31 & 85.00 & 65.61 & 56.35 & 50.84 & 61.71 & 86.01 & 65.58 & 0.00\\
            DMN-ZS (local) \cite{dmn}
                & 42.79 & 47.27 & 60.46 & 50.46 & 75.09 & 46.00 & 72.23 & 77.39 & 73.58 & 77.74 & 80.60 & 73.54 & 87.13 & 85.74 & 68.21 & 58.54 & 54.12 & 63.29 & 86.87 & 67.42 & +1.84\\
            DMN-ZS (global) \cite{dmn}
                & 35.60 & 40.56 & 52.05 & 42.85 & 72.10 & 44.20 & 68.93 & 74.92 & 69.25 & 76.22 & 80.01 & 72.24 & 86.36 & 84.62 & 63.02 & 57.42 & 50.56 & 58.44 & 84.80 & 63.90 & -1.68\\
            {\algname}
                & 43.72 & 48.46 & 61.67 & 51.21 & 76.19 & 46.86 & 73.10 & 78.75 & 74.69 & 78.37 & 81.22 & 73.67 & 87.38 & 86.51 & 68.43 & 60.22 & 55.76 & 63.67 & 87.29 & 68.27 & \textbf{+2.69} \\
            \midrule \midrule
            & \multicolumn{20}{c}{CIFAR-100-C} \\
            \cmidrule(lr){2-22}
            Method & \multicolumn{4}{c}{Noise} & \multicolumn{5}{c}{Blur} & \multicolumn{5}{c}{Weather} & \multicolumn{5}{c}{Digital} & \multirow{2.5}{*}{Total} & \multirow{2.5}{*}{\cellcolor{White}Gain} \\
            \cmidrule(lr){2-5} \cmidrule(lr){6-10} \cmidrule(lr){11-15} \cmidrule(lr){16-20}
            & Gauss. & Shot & Impulse & Speckle 
            & Defocus & Glass & Motion & Zoom & Gauss. 
            & Snow & Frost & Fog & Bright & Spatter
            & Contrast & Elastic & Pixel & JPEG & Saturate \\
            \midrule
            CLIP (ViT-B/16) \cite{clip} 
                & 22.56 & 23.97 & 30.16 & 24.91 & 44.03 & 20.25 & 43.53 & 48.54 & 42.67 & 49.52 & 51.08 & 41.93 & 57.94 & 57.61 & 35.31 & 29.63 & 26.09 & 33.45 & 56.06 & 38.91 & - \\
            VTE \cite{vte}            
                & 19.15 & 20.54 & 27.97 & 21.37 & 40.53 & 18.34 & 39.53 & 43.59 & 37.76 & 47.44 & 48.74 & 42.10 & 54.38 & 55.60 & 49.78 & 30.21 & 29.43 & 30.48 & 51.73 & 37.30 & -1.61\\
            TPT \cite{tpt}
                & 17.13 & 19.02 & 25.62 & 19.37 & 43.33 & 20.12 & 42.89 & 47.98 & 41.27 & 49.72 & 50.87 & 42.48 & 57.19 & 57.75 & 37.87 & 30.52 & 24.93 & 31.92 & 54.00 & 37.58 & -1.33\\
            Zero \cite{zero}
                & 19.23 & 20.82 & 28.90 & 21.65 & 40.98 & 18.27 & 39.81 & 44.07 & 38.04 & 47.80 & 48.67 & 42.61 & 54.57 & 55.67 & 49.64 & 30.34 & 29.74 & 30.74 & 52.01 & 37.55 & -1.36\\
            TDA (local) \cite{tda} 
                & 22.39 & 24.09 & 30.17 & 24.83 & 44.31 & 20.27 & 43.95 & 48.94 & 42.89 & 49.54 & 51.23 & 41.98 & 58.21 & 58.06 & 35.35 & 29.73 & 26.17 & 33.51 & 56.40 & 39.05 & +0.14\\
            TDA (global) \cite{tda} 
                & 22.65 & 24.33 & 30.21 & 24.98 & 44.60 & 19.77 & 43.68 & 48.81 & 42.20 & 49.36 & 51.31 & 41.90 & 58.28 & 57.88 & 35.36 & 29.61 & 26.21 & 33.44 & 56.28 & 38.99 & +0.08\\
            DMN-ZS (local) \cite{dmn}
                & 23.00 & 24.36 & 30.88 & 25.15 & 44.34 & 20.51 & 43.68 & 48.82 & 43.01 & 49.63 & 51.14 & 42.08 & 58.28 & 57.87 & 35.38 & 29.89 & 26.37 & 33.53 & 56.31 & 39.17 & +0.26\\
            DMN-ZS (global) \cite{dmn} 
                & 18.85 & 20.38 & 25.32 & 21.42 & 42.39 & 16.17 & 42.96 & 47.01 & 39.83 & 47.51 & 49.76 & 40.15 & 59.07 & 57.09 & 30.64 & 26.78 & 23.15 & 31.29 & 56.35 & 36.64 & -2.27\\
            {\algname} 
                & 23.34 & 24.74 & 31.36 & 25.26 & 44.97 & 21.08 & 44.03 & 49.58 & 43.32 & 49.66 & 51.45 & 41.99 & 58.64 & 58.51 & 35.20 & 30.16 & 26.78 & 33.80 & 56.87 & 39.51 & \textbf{+0.60} \\
            \bottomrule 
        \end{tabular}
    }
    }
    \caption{Accuracy (mean \% over five random seeds) on corruption benchmarks. Experiments with RN50 backbone is shown in Table \ref{tab:acc:corruption_rn}. }
    \label{tab:acc:corruption_vit}
\end{table*}

\paragraph{Main results (RQ1)}
The experimental results on domain adaptation benchmarks and corruption benchmarks are shown in Table \ref{tab:acc:domain} and \ref{tab:acc:corruption_vit}, respectively. Since each client has limited amount of local data due to data decentralization, the performance of memory-based baselines are constrained, often less effective than the best episodic TTA method. The simple global memory sharing strategy shows improvements on TerraIncognita, but harm the performance on VLCS and corruption benchmarks, even introducing negative transfer, where performance drops below that of the original CLIP model. This suggests that when domain differences are large, such a naive strategy fails to provide adequate personalization for each client. In contrast, {\algname}’s hybrid approach, which combines local and external memory, enables it to effectively integrate information from other clients while preserving personalization, allowing it to outperform both memory-based and episodic TTA baselines. 

\begin{figure}
    \centering
    \includegraphics[width=0.9\linewidth]{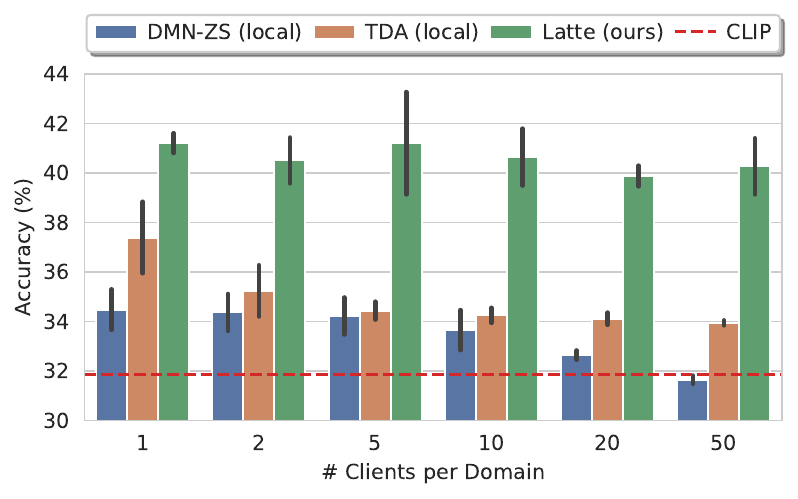}
    \caption{Comparison of memory-based TTA methods under different levels of data decentralization on TerraIncognita.}
    \label{fig:num_clients}
\end{figure}

\paragraph{Different levels of decentralization (RQ1)}
We further analyze the impact of data decentralization on memory-based methods. As shown in Figure \ref{fig:num_clients}, when the number of clients per domain increases, each client has less local data, the performance of DMN-ZS and TDA significantly declines. In contrast, {\algname} remains highly stable, showing minimal sensitivity to data decentralization, demonstrating its robustness in federated settings. 


\begin{table}
    \centering
    \scriptsize
    \resizebox{1.0\linewidth}{!}{
    \setlength{\tabcolsep}{1.0mm}{
        \begin{tabular}{lcccccc}
            \toprule
            \multirow{2.5}{*}{Method} & \multicolumn{5}{c}{Computation} & Communication  \\
            \cmidrule(lr){2-6} \cmidrule(lr){7-7} 
            & Augment & BackProp & MACs & Runtime & Memory & Bytes \\
            \midrule 
            CLIP            & \no   & \no   & 17.6G & 8.3ms     & 1.5G & (172M) \\
            \midrule
            TPT             & \yes  & \yes  & 1.72T & 143.2ms   & 3.6G  & -   \\
            Zero            & \yes  & \no   & 1.13T & 66.6ms    & 2.3G  & -   \\
            DMN-ZS (local)  & \no   & \no   & 17.6G & 8.5ms     & 1.7G  & -  \\
            DMS-ZS (global) & \no   & \no   & 17.6G & 8.5ms + RTT    & 1.7G    & 2.05M  \\
            {\algname}      & \no   & \no   & 17.6G & 8.6ms     & 1.7G  & 614K \\
            \bottomrule  
        \end{tabular}
    }
    }
    \caption{Comparison of communication and computation cost on CIFAR-100-C, where {\algname} has the largest memory size. RTT denotes the round-trip time of communication with the server.}
    \label{tab:efficiency}
\end{table}

\paragraph{Computation cost (RQ2)} 
In Table \ref{tab:efficiency}, we summarize whether each algorithm requires data augmentation (Augment), backpropagation (BackProp), and report the number of multiply–accumulate operations (MACs), runtime, and memory usage to evaluate computation cost. Episodic TTA methods (TPT, Zero) require model inference on multiple augmentations, leading to a significant increase in computation compared to CLIP. Additionally, TPT introduces extra overhead due to prompt tuning. In contrast, {\algname}, like other memory-based methods (e.g., DMN-ZS), achieves high computation efficiency. It only involves lightweight matrix multiplications, adding just 871K MACs, which is negligible compared to the 17.6G MACs required for CLIP's visual encoder inference. 

\begin{figure}
    \centering
    \includegraphics[width=0.9\linewidth]{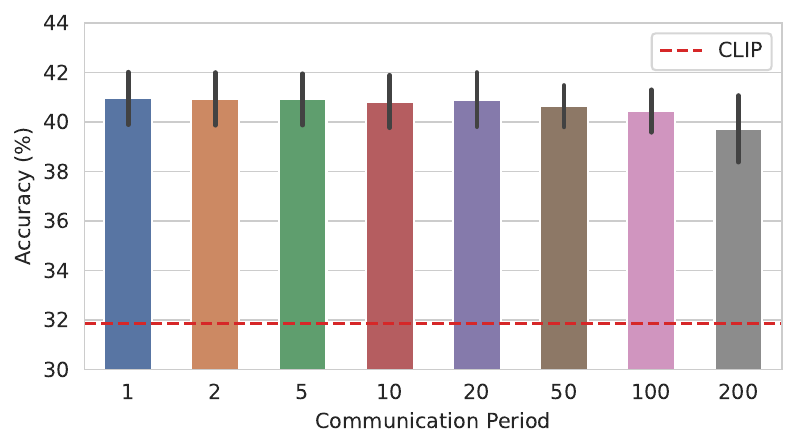}
    \caption{Performance of {\algname} with different communication period on TerraIncognita.}
    \label{fig:comm_period}
\end{figure}

\paragraph{Communication cost (RQ2)}
We measure the data transmission (in bytes) per communication round in {\algname} and compare it to the CLIP visual encoder (which is transmitted only once for model deployment). The per-round communication cost of {\algname} is less than $0.4\%$ of the CLIP visual encoder's size, ensuring high communication efficiency. 
Regarding communication frequency, since {\algname} decouples TTA from communication, it does not require communication for every test sample. In Figure \ref{fig:comm_period}, we test {\algname} by varying the update frequency of external memory, where each client updates it only after observing $T$ samples (with $T$ ranging from 1 to 200). We find that when $T \leq 50$, accuracy remains almost unchanged, and even when $T = 200$, the performance only degrades slightly. Notably, each client in TerraIncognita has $~600$ samples in average, meaning that with $T=200$, each client updates its external memory only 3 to 4 times throughout the entire process. This further validates the communication efficiency of {\algname}. 


\begin{figure}
    \centering
    \includegraphics[width=0.9\linewidth]{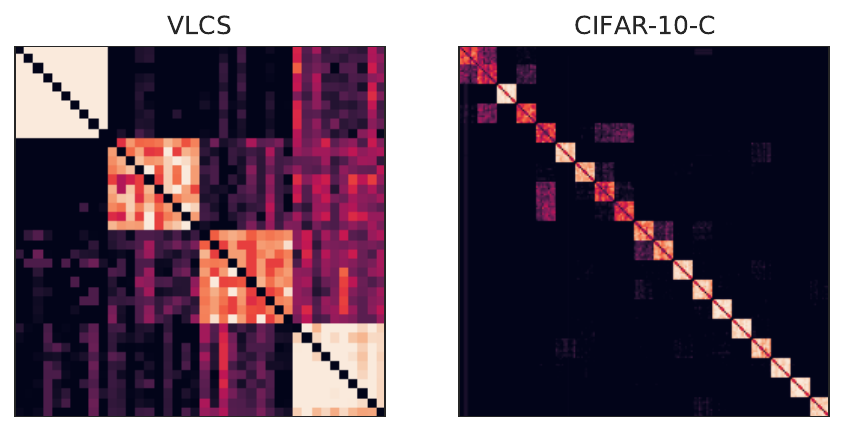}
    \caption{Visualization of the collaboration.  Clients are sorted by domains or corruption types in the same order as in Tables \ref{tab:acc:domain}, \ref{tab:acc:corruption_vit}.}
    \label{fig:collab_vis}
\end{figure}

\begin{figure}
    \centering
    \includegraphics[width=1.0\linewidth]{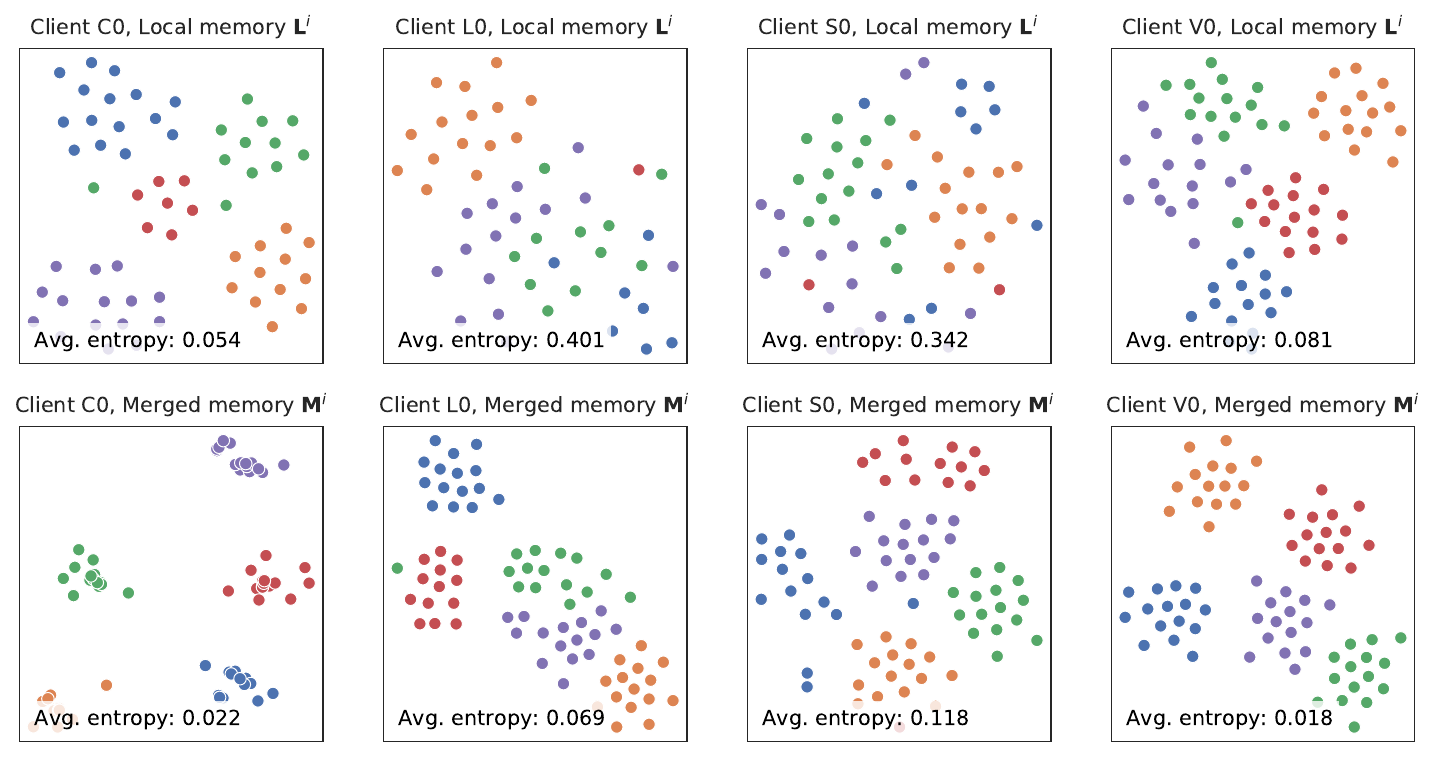}
    \caption{t-SNE \cite{tsne} visualization of local and merged memories. }
    \label{fig:embed_vis}
\end{figure}

\begin{figure}
    \centering
    \includegraphics[width=1.0\linewidth]{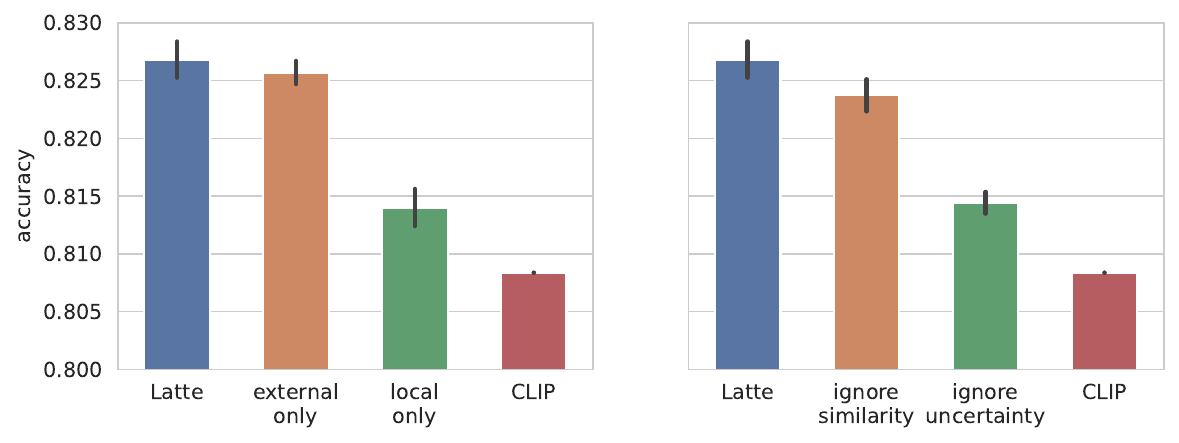}
    \caption{Ablation study on VLCS. }
    \label{fig:ablation}
\end{figure}

\paragraph{Visualization of memory sharing (RQ3)}
Figure \ref{fig:collab_vis} visualizes client collaboration in our VLCS and CIFAR-10-C experiments, where pixel $(i, j)$ represents the frequency at which client $i$ downloads prototypes uploaded by client $j$. Clients are sorted by domains or corruption types in the same order as in Tables \ref{tab:acc:domain} and \ref{tab:acc:corruption_vit} (e.g., in VLCS, the order is C0 - C9, L0 - L9, S0 - S9, V0 - V9). We oberver that clients in {\algname} primarily obtain prototypes from clients with similar distributions, indicating a preference for in-distribution knowledge transfer. 
Furthermore, in Figure \ref{fig:embed_vis}, we select one client per domain and visualize both its local memory $\tL^i$ and merged memory $\tM^i$ after updating with 300 test samples. We observe that local memory, constructed solely from the client's own data, exhibits higher average entropy and a more scattered distribution. In contrast, the merged memory, which incorporates prototypes from clients with similar distributions, shows a significant entropy reduction, with each pseudo-class becoming more clustered, resulting in more representative visual embeddings/prototypes. This improvement in memory quality translates directly into better performance, as verified in Figure \ref{fig:ablation}. 

\paragraph{Ablation study}
During adaptation, {\algname} utilizes the merged memory, which integrates both external and local memories. In Figure \ref{fig:ablation} (left), we compare {\algname} with its variants that use only external memory or only local memory and find that both components contribute to performance improvement. Additionally, {\algname} computes memory logits by considering both similarity $\vf^\top \vm_{y, \kappa}^i$ and uncertainty $H(\vm_{y, \kappa}^i)$. In Figure \ref{fig:ablation} (right), we further analyze their impact and observe that both factors play a crucial role in enhancing performance. 

\begin{figure}
    \centering
    \includegraphics[width=1.0\linewidth]{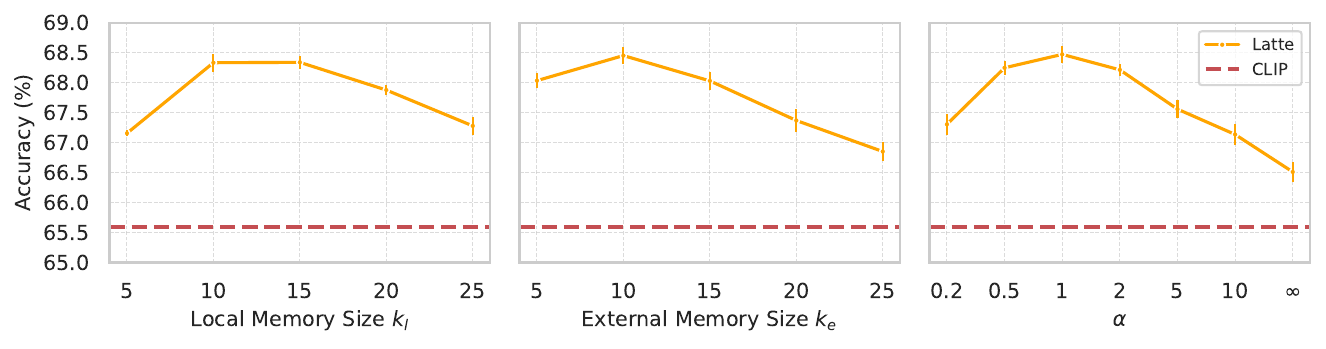}
    \caption{Hyperparameter sensitivity analysis on CIFAR-10-C. }
    \label{fig:hyperparameter_1}
\end{figure}

\paragraph{Hyperparameter sensitivity}
Figure \ref{fig:hyperparameter_1} presents {\algname}'s performance under different local memory size $k_l$, external memory size $k_e$, and $\alpha$ values. While the results for $\beta$ and $\gamma$ are also provided in Table \ref{fig:hyperparameter_2}. We observe that {\algname} consistently improves CLIP's performance across a wide range of hyperparameters, demonstrating its robustness to parameter selection. 

\section{Conclusion}

In this work, we proposed {\algname}, a collaborative test-time adaptation algorithm of vision-language models in federated learning. It includes a memory sharing strategy enabling collaboration and personalization, and an adaptation strategy incorporating both sample similarity and uncertainty. Extensive experiments show that {\algname} significantly improves the model performance, with high efficiency in both computation and communication. 

\section*{Acknowledgement}

This work is supported by National Science Foundation under Award No. IIS-2416070, IIS-2117902. The views and conclusions are those of the authors and should not be interpreted as representing the official policies of the funding agencies or the government.

{
    \small
    \bibliographystyle{ieeenat_fullname}
    \bibliography{main}
}

\appendix
\onecolumn

\section*{Appendix}

\vspace{20pt}

\addtocontents{toc}{\protect\setcounter{tocdepth}{2}}

\tableofcontents

\vspace{20pt}

\section{More Discussion}
\label{appendix:discussion}

\subsection{Additional Related Works}
\label{appendix:discussion:related_works}

\paragraph{Federated learning}
Federated learning is a distributed machine learning system where multiple clients collaborate to train machine learning models under the orchestration of a central server, without sharing their raw data. Many methods focus on fine-tuning VLMs to FL clients. For example, PromptFL \cite{promptfl} applies FedAvg \cite{fedavg} to perform prompt tuning on CLIP, while pFedPrompt \cite{pfedprompt} extends this by introducing personalization. FedCLIP \cite{fedclip} applies an attention-based adapter on image embeddings for fine-tuning. FedTPG \cite{fedtpg} aims to generalize to new classes, learning a unified prompt generation network across multiple remote clients in a scalable manner. PromptFolio \cite{promptfolio} provides a theoretical framework for prompt-based federated learning using feature learning theory. FF-DVP \cite{ff-dvp} explores fair FL frameworks by addressing bias in pre-trained VLMs. However, all these methods assume that clients have labeled data, which is not the case in test-time adaptation.

\paragraph{Test-time adaptation}
Test-time adaptation is a domain adaptation paradigm, adapting a source model to a target domain, without accessing the data on source domain. Before applying to CLIP, entropy minimization and memory are already used in test-time adaptation for more smaller models. For example, Tent \cite{tent} use entropy minimization to update the BatchNorm parameters. Similar to TPT \cite{tpt} and TPS \cite{tps}, MEMO \cite{memo} make multiple augmentation of the input image, and minimize the marginal entropy to enhance the model performance. These algorithms have largely inspired the development of TTA methods tailored for VLMs.












\newpage
\section{Details of Proposed Method}

\subsection{Notation} \label{appendix:method:notation}

For readability, we summarize the notations here. In our paper, 3D tensors are denoted by bold sans-serif uppercase symbols (e.g. $\tT$), matrices are denoted by bold italic uppercase symbols (e.g. $\mM$), vectors are denoted by bold italic lowercase symbols (e.g. $\vv$), and scalars are denoted as math italic lowercase symbols (e.g. $s$). We use superscripts to denote the affiliation of the tensor/matirx/vector, such as $^i$ representing parameters on client $i$; we use 

\begin{table*}[h!]
    \centering
    \small
        \begin{tabular}{rll}
            \toprule
            \multicolumn{2}{l}{\ Notation} & Definition \\
            \midrule
            
            $d$     & & dimension of feature ($d=512$ for ViT-B/16, $d=1,024$ for RN50) \\
            $c$     & & number of classes \\
            $n$     & & number of clients, size of global memory \\
            $k_l$   & & size of local memory \\
            $k_e$   & & size of external memory \\
            
            $\vf$ & $\in \R^d$              & normalized embedding of a new testing image \\
            $\mT$ & $\in \R^{c \times d}$   & normalized text embedding \\
            $\vt_y$ & $\in \R^{d}$          & normalized text embedding of class y \\
            
            $\tL^i$ & $\in \R^{c \times k_l \times d}$  & local memory of client $i$, $\tL^i = \{\mL_1^i, \cdots, \mL_c^i\}$ \\
            $\mL_y^i$ & $ \in \R^{k_l \times d}$         & local memory of class $y$, client $i$, $\mL_y = [\vl_{y, 1}, \cdots, \vl_{y, k_l}]^\top$ \\
            $\vl_{y, \kappa}$ & $ \in \R^{d}$            & $\kappa$-th embedding in local memory of class $y$, client $i$ \\

            $\tG$ & $ \in \R^{c \times n \times d}$  & global memory, $\tG = \{ \mG_1, \cdots, \mG_c \}$ \\
            $\mG_y$ & $ \in \R^{n \times d}$         & global memory of class $y$, $\mG_y = [\vg_{y, 1}, \cdots, \vg_{y, n}]^\top$ \\
            $\vg_{y, i}$ & $ \in \R^{d}$             & $i$-th prototype in global memory of class $y$, uploaded by client $i$ \\
            
            $\tE^i$ & $ \in \R^{c \times k_e \times d}$  & external memory for client $i$, $\tE^i = \{\mE_1^i, \cdots, \mE_c^i\}$ \\
            $\mE_y^i$ & $ \in \R^{k_e \times d}$         & external memory of class $y$, client $i$, $\mE_y = [\ve_{y, 1}, \cdots, \ve_{y, k_l}]^\top$ \\
            $\ve_{y, \kappa}$ & $ \in \R^{d}$            & $\kappa$-th prototype in external memory of class $y$, client $i$ \\

            $H(\cdot)$  & $:\R^d \to \R_+$ & entropy, $H(\vf) = - \sum_{y = 1}^c \hat{p}_{y} \log \hat{p}_y$, where $\hat{p}_y = \frac{\exp(100 \cdot \vf^\top \vt_y)}{\sum_{y'=1}^c \exp(100 \cdot \vf^\top \vt_{y'})}$ \\
            
            \bottomrule 
        \end{tabular}
    \caption{Notation}
    \label{tab:notation}
\end{table*}

\subsection{Pseudo Code} \label{appendix:method:pseudo_code}

\newcommand{\SUB}[1]{\ENSURE \hspace{-0.2in} #1}
\newcommand{\CMT}[1]{\textit{\textcolor{gray}{\# #1}}}
\renewcommand{\algorithmicensure}{}

\begin{figure}[h!]
\centering
\begin{minipage}{0.8\textwidth}
    \begin{algorithm}[H]
    \caption{\algname} \label{alg:main}
    \small
    \begin{algorithmic}[1]
    \SUB{\texttt{Latte}}(client $i$, image $\tI$) \hfill \CMT{Run on client $i$}
    \STATE{\CMT{Step 1 (Subection \ref{subsec:prelim})}}
    \STATE $\vf = \normalize(\mathrm{image\_encoder}(\tI))$ \hfill \CMT{Encode input image into embedding}
    \STATE $\vz_\pre = 100 \cdot \vf^\top \mT^\top$, $\hat{y} = \argmax \vz_\pre$ \hfill \CMT{Get logits and initial prediction}

    \STATE{\CMT{Step 2 (Subection \ref{subsec:local_memory})}}
    \STATE Update local memory $\mL^i_{\hat{y}}$

    \STATE{\CMT{Step 3 (Subection \ref{subsec:adapt})}}
    \STATE $\mM^i_y = \left \{\vm \in \mL^i_y \cup \mE^i_y : H(\vm) \leq \tau(k_l) \right\}, \forall y$ \hfill \CMT{Merge memory}
    \STATE $\vc^i_y = \normalize \left( \sum_{\kappa=1}^k w^i_{y, \kappa}\cdot \vm^i_{y, \kappa} \right) , \forall y$, where \\ $w^i_{y, \kappa} = \exp \left( \beta \cdot \vf^\top \vm^i_{y, \kappa}\right) \cdot \exp \left( - \gamma \cdot H \left(\vm^i_{y, \kappa} \right) \right)$ \hfill \CMT{Compute class prototype}
    \STATE $\vz_{\text{mem}} = 100 \cdot \vf^\top \mC^i$, where $\mC^i = \left[\vc^i_1, \cdots, \vc^i_c\right]$ \hfill \CMT{Compute memory logits}
    \STATE $\vz_{\post} = \vz_{\pre} + \alpha \cdot \vz_{\text{mem}}$, $\hat{y}_{\post} = \argmax \vz_{\post}$ \hfill \CMT{Compute adapted logits and final prediction}

    \STATE{\CMT{Step 4 (Subection \ref{subsec:external_memory})}}
    \IF{connected to the server and ready to communicate}
        \STATE Compute and upload prototypes according to Equation \ref{eq:global_memory}
        \STATE (Wait for the server to retrieve relevant prototypes)
        \STATE Download relevant prototypes and save as external memory $\tE^i$
    \ENDIF
    \end{algorithmic}
    \end{algorithm}
\end{minipage}
\end{figure}

\newpage
\section{Proofs} \label{appendix:proof}

\subsection{Assumptions} \label{appendix:proof:assumption}

We first restate the assumptions and give detailed explanation. 

\begin{assumption}[Data distribution] \label{assumption:data}
    In binary classification setting, with image embeddings in $\vx \in \R^d$ and label $y \in \{0, 1\}$, we assume 
    \begin{enumerate}
        \item Label distribution: The prior label distribution $\Pr(y=0) = \Pr(y=1) = \frac{1}{2}$, 
        \item Conditional feature distribution: for each class $y$, the image embedding is uniformly sampled from a $d$-dimensional unit hypersphere $\sS_y = \{\vf: \| \vf - (2y -1) \vmu \|_2 \leq 1\}$, with center $(2y - 1) \vmu$ (i.e., $\pm \vmu$) and radius 1. 
    \end{enumerate}
\end{assumption}

\begin{remark}
    For clarity, we consider a simple, balanced binary classification problem, though the approach can be extended to multi-class classification, where entropy computation becomes more complex. 
    The conditional feature distribution is adapted from \cite{grad_confuse}, a common assumption in theoretical analysis. We use it to quantify the relationship between decision boundary shifts and error rate changes. This assumption can also be extended to a compact set with bounded density. The irreducible error depends on the relationship between the class center distance and the radius: 
    \begin{itemize}
        \item When $\| \vmu \|_2 \geq 1$, the two classes are linearly separable, 
        \item When $\| \vmu \|_2 < 1$, the two classes overlap and cannot be perfectly classified. 
    \end{itemize}
\end{remark}

\begin{assumption}[Initial classifier] \label{assumption:initial_classifier}
    We assume the initial classifier is a linear classifier, 
    \begin{align}
        z = \vf^\top \vwpre + \bpre, 
    \end{align}
    where $\vwpre, \bpre$ are the weight and bias of the linear layer, with $\| \vwpre \|_2 = 1$. The prediction probability and scalar prediction are defined as
    \begin{align}
        \hat{p}_{1} = \frac{1}{1 + \exp(-tz)}, \quad \hat{p}_0 = 1 - p_1, && \hat{y} = 1\{z > 0\}, 
    \end{align}
    where $t > 0$ is the scaling factor, relevant to the learned temperature parameter of CLIP model, and the entropy is defined as, 
    \begin{align}
         H(\vf) = - \hat{p}_{1} \log \hat{p}_{1} - \hat{p}_{0} \log \hat{p}_{0}
    \end{align}
\end{assumption}

\begin{remark}
    The initial CLIP classifier $100 \cdot \vf^\top \mT^\top$ is a linear layer, and can be transformed to the form in Assumption \ref{assumption:initial_classifier}. Notice that in practice, the CLIP model has no bias term $\bpre$, while the data is not centralized (i.e., the mean of the two class centers is not at the origin). However, for clarity, we assume a centralized setting where the class centers are $\pm \vmu$. To account for this, we allow the initial linear model to include a bias term $\bpre$. Note that these two formulations are equivalent and can be converted by changing the coordinate system. 
\end{remark}

\begin{assumption}[Quality of initial classifier] \label{assumption:initial_classifier_quality}
    We assume the initial classifier's error rate satisfies:
    \begin{align}
        \epsilon_\pre < \frac{1}{2}. 
    \end{align}
    Equivalently, this means
    \begin{align}
        \vmu^\top \vwpre > 0, \quad - \vmu^\top \vwpre - 1 < \bpre < \vmu^\top \vwpre + 1. 
    \end{align}
\end{assumption}

\begin{remark}
    We assume that the initial classifier is not arbitrarily poor, which helps avoid two adversarial cases that TTA inherently cannot handle:
    \begin{itemize}
        \item Label Swap Ambiguity: Since TTA has no access to ground-truth labels, it cannot distinguish between a target distribution and its label-swapped counterpart (i.e., all 1s flipped to 0s and vice versa). As a result, TTA will produce the same adaptation outcome in both cases. This implies that if TTA reduces the error rate in one scenario, it must increase it in the swapped case, proving that TTA cannot universally reduce error rates across all possible target distributions. To prevent this issue, we assume such label swaps do not occur by assuming $\vmu^\top \vwpre > 0$, meaning the initial classifier must be at least somewhat aligned with the true data patterns.
        \item Complete Misclassification of a Class: If the initial classifier assigns the support set of one class to the other class, pseudo-label-based methods will never obtain any correct samples for that class, even as the number of samples approaches infinity. This makes adaptation impossible. To avoid this, we assume the error rate $\epsilon_\pre < \frac{1}{2}$, which implies that the initial classifier does not entirely collapse one class into another.
    \end{itemize}
\end{remark}

\newpage
\subsection{Preliminaries}

In this subsection, we summarize key lemmas related to volumes, which will be used in our subsequent proofs. 

\begin{lemma}[Volume of sphere]
    The volume of a $d$-dimensional unit ball, whose radius $R = 1$, is
    \begin{align}
        \Vs(d) = \frac{\pi^{\frac{d}{2}}}{\Gamma(\frac{d}{2} + 1)}, 
    \end{align}
    where $\Gamma(x) = \int_0^{+\infty} t^{x-1} e^{-t} \dif t$ is the Gamma function. 
\end{lemma}

\begin{lemma}[Volume of sphere cap]
    The volume of a $d$-dimensional sphere cap with radius 1 and polar angle $\theta$ is
    \begin{align}
        \Vsc(d, \theta) = \frac{\pi^{\frac{d-1}{2}}}{\Gamma(\frac{d + 1}{2})} \int_0^{\theta} \sin^d(\varphi) \dif \varphi. 
    \end{align}
\end{lemma}

\begin{corollary}[Lower and upper bounds of volume ratio of sphere cap and sphere] \label{crl:vol_ratio}
    When $\theta \in [0, \frac{\pi}{2}]$, the volume ratio of a $d$-dimensional sphere cap (radius 1, polar angle $\theta$) and a $d$-dimensional unit hypersphere has the following lower and upper bounds
    \begin{align}
        \frac{1}{\sqrt{\pi}} \cdot \frac{1}{\sqrt{2d + 4}} \cdot \left(\frac{2}{\pi}\right)^d \cdot \theta^{d+1} \leq \frac{\Vsc(d, \theta)}{\Vs(d)}
        \leq \frac{1}{\sqrt{\pi}} \cdot \frac{1}{\sqrt{2d + 2}} \cdot \theta^{d + 1}
    \end{align}
\end{corollary}
\begin{proof}
    \begin{align*}
         \frac{\Vsc(d, \theta)}{\Vs(d)} = \frac{1}{\sqrt{\pi}} \cdot \frac{\Gamma(\frac{d}{2} + 1)}{\Gamma(\frac{d + 1}{2})} \cdot \int_0^{\theta} \sin^d(\varphi) \dif \varphi
    \end{align*}
    We first give upper and lower bounds of $\frac{\Gamma(\frac{d}{2} + 1)}{\Gamma(\frac{d + 1}{2})}$. Based on Wendel's inequality \cite{wendel}, for any $x > 0$ and $a \in (0, 1)$, we have
    \begin{align*}
        \left(\frac{x}{x+a}\right)^{1-a} \leq \frac{\Gamma(x + a)}{x^a \Gamma(x)} \leq 1
    \end{align*}
    Let $x = \frac{d+1}{2}$ and $a = \frac{1}{2}$, we get
    \begin{align*}
        \frac{\frac{d+1}{2}}{\sqrt{\frac{d}{2}+1}} \leq \frac{\Gamma(\frac{d}{2} + 1)}{\Gamma(\frac{d + 1}{2})} \leq \sqrt{\frac{d+1}{2}}. 
    \end{align*}
    Then for $\int_0^{\theta} \sin^d(\varphi) \dif \varphi$, notice that for $\theta \in [0, \frac{\pi}{2}]$, we have $\frac{2}{\pi} \cdot \theta \leq \sin(\theta) \leq \theta$, therefore
    \begin{align*}
        \int_0^{\theta} \sin^d(\varphi) \dif \varphi 
        &\geq \int_0^{\theta} \left(\frac{2}{\pi}\varphi\right)^d \dif \varphi 
        = \left(\frac{2}{\pi}\right)^d \cdot \frac{1}{d + 1} \theta^{d + 1},  \\
        \int_0^{\theta} \sin^d(\varphi) \dif \varphi 
        &\leq \int_0^{\theta} \varphi^d \dif \varphi 
        = \frac{1}{d + 1} \theta^{d + 1} 
    \end{align*}
    Combine together, 
    \begin{align*}
        \frac{1}{\sqrt{\pi}} \cdot \frac{1}{\sqrt{2d + 4}} \cdot \left(\frac{2}{\pi}\right)^d \cdot \theta^{d+1} \leq \frac{\Vsc(d, \theta)}{\Vs(d)}
        \leq \frac{1}{\sqrt{\pi}} \cdot \frac{1}{\sqrt{2d + 2}} \cdot \theta^{d + 1}
    \end{align*}
\end{proof}


\newpage
\subsection{Asymptotic Analysis on Single Client} \label{appendix:proof:asym}

In this subsection, we analyze the behavior of {\algname} on single client when the sample size $n \to \infty$. It refers to the scenario where each client independently conducts test-time adaptation, without any communication. For clarity, we omit the superscript $^i$ in this subsection. 

\begin{lemma}[Equivalence to 1-NN] \label{lem:1nn}
    When $\alpha, \beta \to + \infty$, the {\algname} classifier is equivalent to an 1-nearest-neighbor (1-NN) classifier. 
\end{lemma}
\begin{proof}
    For each class $y = 1, \cdots, c$, the saved memory $\mM_y$ contains $k$ vectors $\{\vm_{y, 1}, \cdots, \vm_{y, k_e}\}$. {\algname} first compute the class embedding for each class: 
    \begin{align*}
        w_{y, \kappa} &= \exp \left( \beta \cdot \vf^\top \vm_{y, \kappa}\right) \cdot \exp \left( - \gamma \cdot H \left(\vm_{y, \kappa} \right) \right), \\
        \vc_y &= \normalize \left( \sum_{\kappa=1}^k w_{y, \kappa}\cdot \vm_{y, \kappa} \right). 
    \end{align*}
    When $\beta \to +\infty$, for each two $\vm_{y, \kappa}$ and $\vm_{y, \kappa'}$, if $\vf^\top \vm_{y, \kappa} < \vf^\top \vm_{y, \kappa'}$, then
    \begin{align*}
        \frac{w_{y, \kappa}}{w_{y, \kappa'}} = \exp(\beta \cdot (\vf^\top \vm_{y, \kappa} - \vf^\top \vm_{y, \kappa'})) \cdot \exp(-\gamma \cdot (H(\vm_{y, \kappa}) - H(\vm_{y, \kappa'}))) \xrightarrow{\beta \to +\infty} 0
    \end{align*}
    Assuming there are no ties, the aggregation weights become one-hot when $\beta\to+\infty$, then we have, 
    \begin{align*}
        \vc_y \xrightarrow{\beta \to +\infty} \vm_{y, \kappa^*}, \quad \text{where } \kappa^* = \argmax_{\kappa} \vf \vm_{y, \kappa}^\top. 
    \end{align*}
    When $\alpha \to + \infty$, the final classification will only depend on the memory classifier, 
    \begin{align*}
        \hat{y} = \argmax_y \vf^\top \vc_y. 
    \end{align*}
    This is equivalent to finding the vector in the current memory $\mM_y$ that has the highest cosine similarity with $\vf$ and assigning $\vf$ to its corresponding class. Since both $\vf$ and the vectors in memory have unit length, maximizing cosine similarity is equivalent to minimizing the $L^2$ distance. Therefore, the {\algname} classifier is equivalent to an 1-NN classifier. 
\end{proof}



\begin{lemma}[Asymptotic convergence]
    As $n\to\infty$, the memory will converge to 
    \begin{align}
        \vm_{0, \kappa} \overset{\textnormal P}{\to} \vm_0^* := -\vmu - \vwpre, \quad \vm_{1, \kappa} \overset{\textnormal P}{\to} \vm_1^* := + \vmu + \vwpre, \quad \forall \kappa\in[k].
    \end{align}
\end{lemma}

\begin{proof}
Let $\vf_1,\dots,\vf_n\in\sB(+\vmu, 1)\cup\sB(-\vmu, 1)$ denote the i.i.d.\ uniform samples observed so far. For a body $\sK\subset\mathbb R^d$, let $\Vol(\sK)$ denote its $d$-dimensional volume. First, consider class 1. For $b\ge0$, let $\sG(\boldsymbol\xi,b):=\{\vf\in\sB(\boldsymbol\xi,1):\vf^\top \vwpre+\bpre>b\}$. Recall that
\begin{align*}
\{\vm_{1,1},\dots,\vm_{1,k}\}=\underset{\vf_i:\,\vf_i^\top \vwpre + \bpre > 0}{\operatorname{arg\,top}_k}(-H(\vf_i))=\underset{\vf_i:\,\vf_i\in R(0)}{\operatorname{arg\,top}_k}(\vf_i^\top \vwpre+\bpre).
\end{align*}
Since $\epsilon_\pre<\frac12$, then (i) $\vm_1^* = +\vmu+\vwpre\in \sG(0)\cap\sB(+\vmu,1)$ is the unique maximizer of the linear function $h(\vf):=\vf^\top\vw+\bpre$ over $\sB(+\vmu, 1) \cup \sB(-\vmu, 1)$, and (ii) $0<\Vol(\sG(b))<\Vol(\sB(+\vmu,1)\cup\sB(-\vmu,1))$ for every $0\le b<h(\vm_1^*)$. 

For any $0<\rho<1$, since $\sG(0)\cap\sB(\vm_1^*,\rho)$ is a strictly convex body, then there exists $0<\rho'<\rho$ such that $\sG(h(\vm_1^*)-\rho')=\sG(h(\vm_1^*-\rho'\vwpre))\subset \sG(0)\cap\sB(\vm_1^*,\rho)$. 
Let
\begin{align*}
p_\rho:=\frac{\Vol(\sG(+\vmu,h(\vm_1^*)-\rho'))+\Vol(\sG(-\vmu,h(\vm_1^*)-\rho'))}{\Vol(\sB(+\vmu,1))+\Vol(\sB(-\vmu,1))}.
\end{align*}Since $0<p_\rho<1$, and $1_{[\vf_i\in R(h(\vm_1^*-\rho'\vwpre))]}\sim\operatorname{Bern}(p_\rho)$, then for any $\kappa\in[k]$, as $n\to\infty$,
\begin{align*}
\Pr[\|\vm_{1,\kappa}-\vm_1^*\|>\rho]={}&\Pr[\vm_{1,\kappa}\notin \sG(0)\cap\sB(\vm_1^*,\rho)]
\\\le{}&\Pr[\vm_{1,\kappa}\notin\sG(h(\vm_1^*)-\rho')]
\\\le{}&\Pr[\exists\kappa\in[k]:\,\vm_{1,\kappa}\notin\sG(h(\vm_1^*)-\rho')]
\\={}&\Pr\Big[\underset{\vf_i:\,\vf_i\in\sG(0)}{\operatorname{arg\,top}_k}h(\vf_i)\not\subset\sG(h(\vm_1^*)-\rho')\Big]
\\={}&\Pr\bigg[\sum_{i\in[n]}1_{[\vf_i\in\sG(h(\vm_1^*)-\rho')]}<k\bigg]
\\={}&\sum_{j=0}^{k-1}\binom njp_\rho^j(1-p_\rho)^{n-j}
\\={}&(1-p_\rho)^n\sum_{j=0}^{k-1}\binom njp_\rho^j(1-p_\rho)^{-j}
\\={}&(1-p_\rho)^n\sum_{j=0}^{k-1}\gO(n^j)
\\={}&(1-p_\rho)^n\cdot \gO(n^{k-1})\to 0.
\end{align*}
This implies that $\vm_{1, \kappa} \overset{\textnormal P}{\to} \vm_1^*$ for every $\kappa\in[k]$. Similarly, we can show that $\vm_{0, \kappa} \overset{\textnormal P}{\to} \vm_0^*$ for every $\kappa\in[k]$.
\end{proof}


\begin{lemma}[Better classifier]
    When $\vm_{0, \kappa} = \vm_0^*$ and $\vm_{1, \kappa} = \vm_1^*$ for all $\kappa = 1, \cdots, k$, the converged classifier is guaranteed to be better than the initial classifier, with
    \begin{align}
        \vmu^\top \vwasym \geq \vmu^\top \vwpre
    \end{align}
    and the error rate is lower: $\epsilon_{\asym} \leq \epsilon_{\pre}$.
\end{lemma}
\begin{remark}
    Intuitively, it means that when the TTA algorithm converge, the post-TTA classifier is better than the pre-TTA classifier. 
\end{remark}

\begin{proof}
    The pre-TTA classifier is $\vwpre, \bpre$. the post-TTA classifier is
    \begin{align*}
        \vwasym = \frac{\vmu + \vwpre}{\| \vmu + \vwpre \|_2}, \quad \bpost = 0, 
    \end{align*}
    whose decision boundary is the perpendicular bisector of $\vm_0^*$ and $\vm_1^*$. Notice that
    \begin{align*}
        \vmu^\top \vwasym \geq \vmu^\top \vwpre
    \end{align*}
    To prove this, 
    \begin{align*}
        &\quad\ \ \vmu^\top \vwpost \geq \vmu^\top \vwpre \\
        & \Leftrightarrow \vmu^\top \vmu + \vmu^\top \vwpre \geq \| \vmu + \vwpre \|_2 \cdot (\vmu^\top \vwpre) \\
        & \Leftrightarrow (\vmu^\top \vmu + \vmu^\top \vwpre)^2 \geq \| \vmu + \vwpre \|_2^2 \cdot (\vmu^\top \vwpre)^2 \\
        & \Leftrightarrow (\vmu^\top \vmu + 2 \vmu^\top \vwpre) (\vmu^\top \vmu - (\vmu^\top \vwpre)^2 ) \geq 0
    \end{align*}
    Notice that $\vmu^\top \vmu + 2 \vmu^\top \vwpre > 0$, and $\vmu^\top \vmu = \| \vmu \|_2^2 \geq (\vmu^\top \vwpre)^2$ since $\| \vwpre \|_2 = 1$. So this is proved. 

    This also leads to lower error rate. For a linear classifier with parameter $\vw$ and $b = 0$, the error rate is
    \begin{align*}
        \epsilon = \begin{cases}
            0, & \vmu^\top \vw > 1 \\
            \frac{\Vsc(\arccos(\vmu \vw^\top))}{\Vs}, & 0 < \vmu^\top \vw \leq 1
        \end{cases}
    \end{align*}
    which monotonously decreases when $\vmu^\top \vw $ increases. So we can prove that
    \begin{align*}
        \epsilon_\asym := \epsilon(\vwasym, \basym) = \epsilon(\vwasym, 0) \leq \epsilon(\vwpre, 0) \leq \epsilon(\vwpre, \bpre) =: \epsilon_\pre
    \end{align*}
\end{proof}

\newpage
\subsection{Non-Asymptotic Analysis on Single Client} \label{appendix:proof:nonasym}

\begin{lemma}[Convergence of embeddings in memory] \label{lem:embedding}
    With probability at least $1 - \delta$, 
    \begin{itemize}
        \item All embeddings in $\mM_0$ distribute in the sphere cap with sphere center $-\vmu$, apex $\vm_0^* = -\vmu -\vwpre$, and polar angle $\theta$; 
        \item All embeddings in $\mM_1$ distribute in the sphere cap with sphere center $+\vmu$, apex $\vm_1^* = +\vmu +\vwpre$, and polar angle $\theta$; 
    \end{itemize}
    where the polar angle is
    \begin{align}
        \theta = \frac{\pi}{2} \left( \frac{8}{\sqrt{\pi}} \cdot \sqrt{2d + 4} \cdot \frac{\log (2 / \delta ) + k}{N} \right)^\frac{1}{d+1}
    \end{align}
    with sufficiently large $N$ such that $\theta < \min\{ \frac{\pi}{12}, \arccos(\max\{0, 1 - 2\vmu^\top \vwpre\}), \arccos(\max\{0, |\bpre| - \vmu^\top \vwpre\})\}$. 
\end{lemma}

\begin{proof}
    We partition the support set of two spheres $\sS_0 \cup \sS_1$ into three non-overlapping regions: 
    \begin{itemize}
        \item Sphere cap $\sC_0$: the sphere cap with sphere center $-\vmu$, apex $\vm_0^* = -\vmu -\vwpre$, and polar angle $\theta$; 
        \item Sphere cap $\sC_1$: the sphere cap with sphere center $+\vmu$, apex $\vm_1^* = +\vmu +\vwpre$, and polar angle $\theta$; 
        \item The rest of the space in the two spheres. 
    \end{itemize}
    When $\theta < \min\{ \arccos(\max\{0, 1 - 2\vmu^\top \vwpre\}), \arccos(\max\{0, |\bpre| - \vmu^\top \vwpre\})\}$, a sufficient condition for our proof is that, among the $N$ seen test sample embeddings, at least $k$ belong to $\sC_0$ and at least $k$ belong to $\sC_1$. Let $N_0$ and $N_1$ denote the number of seen test sample embeddings belonging to $\sC_0$ and $\sC_1$, respectively. For a test sample with embedding $\vf$, suppose $p = \Pr(\vf \in \sC_0) = \Pr(\vf \in \sC_1)$ (by symmetric), then we have 
    \begin{align*}
        N_0 \sim \mathrm{Binomial}(N, p), \quad N_1 \sim \mathrm{Binomial}(N, p), 
    \end{align*}
    while $N_0, N_1$ are not independent. $\E[N_0] = \E[N_1] = Np$. When $k < Np$, the Chernoff Bound gives
    \begin{align*}
        \Pr(N_0 < k) &\leq \Pr(N_0 \leq k) \\
        &= \Pr\left(N_0 \leq \left(1 - \frac{Np - k}{Np} \right) Np \right) \\
        &\leq \exp \left( - \left( \frac{Np - k}{Np} \right)^2 \frac{Np}{2} \right) \tag{Chernoff Bound} \\
        &= \exp\left(- \frac{(Np - k)^2}{2Np} \right)
    \end{align*}
    Similar result holds for $N_1$. Finally
    \begin{align*}
        \Pr(N_0 \geq k, N_1 \geq k) \geq 1 - \Pr(N_0 < k) - \Pr(N_1 < k) \geq 1 - 2\exp\left(- \frac{(Np - k)^2}{2Np} \right)
    \end{align*}
    Define $\delta = 2\exp(- \frac{(Np - k)^2}{2Np})$, we have
    \begin{align*}
        \frac{(Np - k)^2}{2 Np} &= \log \left( \frac{2}{\delta} \right) \\
        Np &= \log \left( \frac{2}{\delta} \right) + k + \sqrt{\left( \log \left( \frac{2}{\delta} \right)\right)^2 + 2 \log \left( \frac{2}{\delta} \right) k} < 2 \left( \log \left( \frac{2}{\delta} \right) + k \right) \\
        p &< \frac{2(\log (2 / \delta ) + k )}{N}
    \end{align*}
    Meanwhile, we have
    \begin{align*}
        p &\geq \frac{\Vsc(d, \theta)}{2 \Vs(d)} \\
        &\geq \frac{1}{2 \sqrt{\pi}} \cdot \frac{1}{\sqrt{2d + 4}} \cdot \left(\frac{2}{\pi}\right)^d \cdot \theta^{d+1} \\
        &= \frac{\sqrt{\pi}}{4} \cdot \frac{1}{\sqrt{2d + 4}} \cdot \left(\frac{2}{\pi} \theta\right)^{d+1}
    \end{align*}
    Put together, we have
    \begin{align*}
        \theta < \frac{\pi}{2} \left( \frac{8}{\sqrt{\pi}} \cdot \sqrt{2d + 4} \cdot \frac{\log (2 / \delta ) + k}{N} \right)^\frac{1}{d+1}. 
    \end{align*}
    which limits to zero when $N \to +\infty$. 

\end{proof}


\begin{theorem}[Convergence of error rate] \label{thm:single_client}
    With probability at least $1 - \delta$, with sufficiently large $N$ such that $\theta < \min\{ \frac{\pi}{12}, \arccos(\max\{0, 1 - 2\vmu^\top \vwpre\}), \arccos(\max\{0, |\bpre| - \vmu^\top \vwpre\})\}$, the error rate satisfies
    \begin{align}
        \gO \left( d^\frac{1}{2} \cdot \left(\frac{\log(2 / \delta) + k}{N}\right)^\frac{1}{d+1} \right), 
    \end{align}
    and when $\epsilon_\asym = 0$, we have a tighter bound of the error rate
    \begin{align}
        \gO \left(d^{- \frac{1}{4}} \cdot C^d \cdot \left(\frac{\log(2 / \delta) + k}{N}\right)^\frac{1}{2} \right), 
    \end{align}
    where $C = \frac{\pi}{2} \cdot \left(\frac{3 \pi}{2} \right)^\frac{1}{2}$. 
\end{theorem}

\begin{proof}
    In Lemma \ref{lem:embedding}, we show that under the same condition, all embeddings in the memory distribute in two sphere caps $\sC_0, \sC_1$ defined in Lemma \ref{lem:embedding}. In this part, we prove that under this condition, the error rate can be correspondingly bounded. 

    
    Denote $h_{\post}(\vf)$ as the classifier after observing $N$ test samples, which is typically not a linear model. We have
    \begin{align*}
        \epsilon(h_\post) 
        &= \frac{1}{2}\Pr_{\vf \sim U(\sS_0)}(h_\post(\vf) = 1) + \frac{1}{2}\Pr_{\vf \sim U(\sS_1)}(h_\post(\vf) = 0) \\
        &= \frac{1}{2} \left( \Pr_{\vf \sim U(\sS_0)}(h_\asym(\vf) = 1, h_\post(\vf) = 1) + \Pr_{\vf \sim U(\sS_0)}(h_\asym(\vf) = 0, h_\post(\vf) = 1) \right) + \\
        &\quad \ \frac{1}{2} \left( \Pr_{\vf \sim U(\sS_1)}(h_\asym(\vf) = 0, h_\post(\vf) = 0) + \Pr_{\vf \sim U(\sS_1)}(h_\asym(\vf) = 1, h_\post(\vf) = 0) \right) \\
        &\leq \frac{1}{2} \left( \Pr_{\vf \sim U(\sS_0)}(h_\asym(\vf) = 1) + \Pr_{\vf \sim U(\sS_0)}(h_\asym(\vf) = 0, h_\post(\vf) = 1) \right) + \\
        &\quad \ \frac{1}{2} \left( \Pr_{\vf \sim U(\sS_1)}(h_\asym(\vf) = 0) + \Pr_{\vf \sim U(\sS_1)}(h_\asym(\vf) = 1, h_\post(\vf) = 0) \right) \\
        &= \epsilon(h_\asym) + \frac{1}{2} \Pr_{\vf \sim U(\sS_0)}(h_\asym(\vf) = 0, h_\post(\vf) = 1) + 
        \frac{1}{2} \Pr_{\vf \sim U(\sS_1)}(h_\asym(\vf) = 1, h_\post(\vf) = 0)
    \end{align*}

    By symmetry, we give an upper bound of $\Pr_{\vf \sim U(\sS_0)}(h_\asym(\vf) = 0, h_\post(\vf) = 1)$ in the rest of the proof. 

    
    For all saved embeddings $\vm_{0, \kappa}$ in $\sC_0$, the distance between $\vm_{0, \kappa}$ and $\vm_0^*$ can be bounded by
    \begin{align*}
        \| \vm_{0, \kappa}  - \vm_0^* \|_2 \leq 2 \sin \frac{\theta}{2} \leq 2 \cdot \frac{\theta}{2} = \theta
    \end{align*}
    Similarly, $\| \vm_{1, \kappa}  - \vm_1^* \|_2 \leq \theta, \forall \kappa$. 

    The exact form of $h_\post(\vf)$ depends on how $\{\vm_{y, \kappa}\}$ are distributed in $\sC_0, \sC_1$, however, we can derive a sufficient condition of $h_\post(\vf) = 0$: For all embedding $\vf$, if $\| \vf - \vm_1^* \|_2 - \| \vf - \vm_0^* \|_2 > 2\theta$, we have
    \begin{align*}
        \min_{\kappa} \| \vf - \vm_{0, \kappa} \|_2 &\leq \min_{\kappa} \left( \| \vf - \vm_0^* \|_2 + \| \vm_0^* - \vm_{0, \kappa} \|_2 \right) \\
        &\leq \| \vf - \vm_0^* \|_2 + \theta \\
        &< \| \vf - \vm_1^* \|_2 - 2\theta + \theta \\ 
        &= \| \vf - \vm_1^* \|_2 - \theta \\ 
        &\leq \| \vf - \vm_{1, \kappa'}\|_2 + \| \vm_{1, \kappa'} - \vm_1^* \|_2   - \theta \\ 
        &\leq \| \vf - \vm_{1, \kappa'}\|_2 + \theta - \theta \\
        &=  \| \vf - \vm_{1, \kappa'}\|_2
    \end{align*}
    Notice that this holds for all $\kappa'$, therefore
    \begin{align*}
        \min_{\kappa} \| \vf - \vm_{0, \kappa} \|_2 < \min_{\kappa'} \| \vf - \vm_{1, \kappa'}\|_2
    \end{align*}
    which indicates $h_\post(\vf) = 0$. Therefore we have
    \begin{align*}
        \Pr_{\vf \sim U(\sS_0)}(h_\asym(\vf) = 0, h_\post(\vf) = 1) 
        \leq \Pr_{\vf \sim U(\sS_0)}(0 \leq \| \vf - \vm_1^* \|_2 - \| \vf - \vm_0^* \|_2 \leq 2\theta)
    \end{align*}

    Exactly computing this probability may still be non-trivial. Therefore, we give a coarse upper bound of it from a geometric perspective. We partition $\sS_0$ into three regions $\sA_1, \sA_2, \sA_3$: 
    \begin{itemize}
        \item $\sA_1 = \{\vf \in \sS_0 : \vf^\top \vwasym > 0\}$: This is the sphere cap where $h_\asym(\vf) = 1$ and we can guarantee $\| \vf - \vm_1^* \|_2 - \| \vf - \vm_0^* \|_2 < 0$. 
        \item $\sA_2 = \{\vf \in \sS_0 : - \Delta \leq \vf^\top \vwasym \leq 0\}$: We construct $\sA_2$ such that $\vf \in \sA_2$ is a necessary condition of $0 \leq \| \vf - \vm_1^* \|_2 - \| \vf - \vm_0^* \|_2 \leq 2\theta$. 
        \item $\sA_3 = \{\vf \in \sS_0: \vf^\top \vwasym < -\Delta\}$: This is the region where we can guarantee $\| \vf - \vm_1^* \|_2 - \| \vf - \vm_0^* \|_2 > 2\theta$
    \end{itemize}
    We have
    \begin{align*}
        \Pr_{\vf \sim U(\sS_0)}(0 \leq \| \vf - \vm_1^* \|_2 - \| \vf - \vm_0^* \|_2 \leq 2\theta) \leq \frac{\Vol(\sA_2)}{\Vol(\sS_0)} = \frac{\Vol(\sA_2)}{\Vs(d)}
    \end{align*}
    where $\Vol$ is the volume function. Finally, we derive a $\Delta$ that can guarantee $\| \vf - \vm_1^* \|_2 - \| \vf - \vm_0^* \|_2 > 2\theta$, and calculate the corresponding $\Vol(\sA_2)$.
    
    Let
    \begin{align*}
        \Pi(\vf) = (\vf^\top \vwasym) \vwasym
    \end{align*}
    be the projection of an embedding $\vf$ to the line connecting $\vm_0^*$ and $\vm_1^*$, and we further define
    \begin{align*}
        A := \| \vm_0^* - \bm{0} \|_2, && B := \max_{\vf \in \sS_0} \| \vf - \Pi (\vf)\|_2, 
    \end{align*}
    and give their lower/upper bound
    \begin{align*}
        A &= \| \vm_0^* - \bm{0} \|_2 = \| \vmu + \vwpre \|_2 > \max\{\| \vwpre \|_2, \| \vmu \|_2\} \geq 1, \\
        B &= \max_{\vf \in \sS_0} \| \vf - \Pi (\vf)\|_2 \leq \max_{\vf \in \sS_0} \| \vf - \vm_0^*\|_2 = 2,  
    \end{align*}
    For any embedding in $\sA_3$, we have
    \begin{align*}
        \| \vf - \vm_1^* \|_2 - \| \vf - \vm_0^* \|_2 
        &= \sqrt{\| \vf - \Pi(\vf) \|_2^2 + \| \Pi(\vf) - \vm_1^* \|_2^2} - \sqrt{\| \vf - \Pi(\vf) \|_2^2 + \| \Pi(\vf) - \vm_0^* \|_2^2} \\
        &\geq \sqrt{B^2 + (A + \Delta)^2} - \sqrt{B^2 + (A - \Delta)^2}
    \end{align*}
    When $\Delta = \theta \sqrt{1 + \frac{4B^2}{4A^2 - \theta^2}}$, we have
    \begin{align*}
        \sqrt{B^2 + (A + \Delta)^2} - \sqrt{B^2 + (A - \Delta)^2} = 2\theta
    \end{align*}
    Notice that when $\theta \leq \frac{\pi}{12}$, we have $ \sqrt{1 + \frac{4B^2}{4A^2 - \theta^2}} < \sqrt{1 + \frac{16}{4 - \left(\frac{\pi}{12}\right)^2}} = 2.251\cdots < 3$, and $\Delta \leq \frac{\pi}{12} \cdot \sqrt{1 + \frac{16}{4 - \left(\frac{\pi}{12}\right)^2}} = 0.589\cdots < 1$. 

    Finally, we give an upper bound of $\Vol(\sA_2)$ to finish our proof. We consider two scenarios: 
    \begin{enumerate}
        \item When $\epsilon_\asym > 0$, $\Vol(\sA_2)$ can be bounded by
        \begin{align*}
            \frac{\Vol(\sA_2)}{\Vs(d)} 
            &< \frac{\Vs(d - 1) \cdot \Delta}{\Vs(d)} \\
            &= \frac{\pi^{\frac{d-1}{2}}}{\Gamma(\frac{d + 1}{2})} \cdot \frac{\Gamma(\frac{d}{2}+1)}{\pi^{\frac{d}{2}}} \cdot \Delta \\
            &\leq \frac{1}{\sqrt{\pi}} \cdot \sqrt{\frac{d+1}{2}} \cdot \Delta \tag{Wendel's inequality and Corollary \ref{crl:vol_ratio}} \\
            &< \frac{1}{\sqrt{\pi}} \cdot \sqrt{\frac{d+1}{2}} \cdot 3 \theta \\
            &\leq \frac{1}{\sqrt{\pi}} \cdot \sqrt{\frac{d+1}{2}} \cdot 3 \cdot \frac{\pi}{2} \left( \frac{8}{\sqrt{\pi}} \cdot \sqrt{2d + 4} \cdot \frac{\log (2 / \delta ) + k}{N} \right)^\frac{1}{d+1} \\
            &\leq \gO \left( d^\frac{1}{2} \cdot \left(\frac{\log(2 / \delta) + k}{N}\right)^\frac{1}{d+1} \right)
        \end{align*}

        \item When $\epsilon_\asym = 0$, we can have a tighter bound
        \begin{align*}
            \frac{\Vol(\sA_2)}{\Vs(d)} 
            &< \frac{\Vsc(d, \arccos(1 - \Delta))}{\Vs(d)} 
        \end{align*}
        Notice that $\arccos(1 - \Delta) \leq \frac{\pi}{2} \cdot \sqrt{\Delta}, \forall \Delta \in [0, 1]$, therefore, 
        \begin{align*}
            \frac{\Vol(\sA_2)}{\Vs(d)} 
            &< \frac{\Vsc(d, \arccos(1 - \Delta))}{\Vs(d)} \\
            &\leq \frac{\Vsc(d, \frac{\pi}{2} \sqrt{\Delta})}{\Vs(d)} \\
            &\leq \frac{1}{\sqrt{\pi}} \cdot \frac{1}{\sqrt{2d + 2}} \cdot \left(\frac{\pi}{2} \sqrt{\Delta}\right)^{d+1} \\
            &\leq \frac{1}{\sqrt{\pi}} \cdot \frac{1}{\sqrt{2d + 2}}  \cdot \left(\frac{\pi}{2} \cdot \left(\frac{3 \pi}{2} \right)^\frac{1}{2} \right)^{d+1} \cdot  \left( \frac{8}{\sqrt{\pi}} \cdot \sqrt{2d + 4} \cdot \frac{\log (2 / \delta ) + k}{N} \right)^\frac{1}{2} \\
            &\leq \gO \left(d^{- \frac{1}{4}} \cdot C^d \cdot \left(\frac{\log(2 / \delta) + k}{N}\right)^\frac{1}{2} \right)
        \end{align*}
        where $C = \frac{\pi}{2} \cdot \left(\frac{3 \pi}{2} \right)^\frac{1}{2}$. 
    \end{enumerate}
\end{proof}

\newpage
\subsection{Non-Asymptotic Analysis on Multiple Clients}

In the final part, we extend of theorem to multiple clients. We consider there are $n$ clients, and we analyze the error rate of {\algname} on a client $i$ among them, which is our ``target'' client. Among these $n$ clients, we assume there are $\nID$ ID clients with the same distribution as client $i$ (including client $i$ itself), and $\nOOD$ clients with distinct distributions. All the assumptions we introduced in Appendix \ref{appendix:proof:assumption} still apply to client $i$ and all ID clients. For OOD clients, we make an additional assumption as follows. 

\begin{assumption}[OOD clients] \label{assumption:ood}
    For each OOD client, we assume its distribution follows all assumptions in Appendix \ref{appendix:proof:assumption}, except the center for two classes' hypersphere are different from $\pm \vmu$. Denote $\vmu_0 = - \vmu$ and $\vmu_1 = + \vmu$. For each OOD clients, if its two classes' hypersphere are centered at $\vmu_0', \vmu_1'$, we assume
    \begin{align}
        \| \vmu_y - \vmu_{y'}' \|_2 > 4, \forall y, y' \in \{0, 1\}
    \end{align}
\end{assumption}

\begin{remark}
    This means that two clients from different domains has non-overlapping and distinct distributions. 
\end{remark}

\begin{theorem}
    Suppose the number of seen testing samples for each client is $N$. With probability at least $1 - \delta$, with sufficiently large $N$ such that $\theta < \min\{ \frac{\pi}{12}, \arccos(\max\{0, 1 - 2\vmu^\top \vwpre\}), \arccos(\max\{0, |\bpre| - \vmu^\top \vwpre\})\}$, the error rate satisfies
    \begin{align}
        \gO \left( d^\frac{1}{2} \cdot \left(\frac{\log(2 / \delta) + k}{\nID \cdot N}\right)^\frac{1}{d+1} \right), 
    \end{align}
    and when $\epsilon_\asym = 0$, we have a tighter bound of the error rate
    \begin{align}
        \gO \left(d^{- \frac{1}{4}} \cdot C^d \cdot \left(\frac{\log(2 / \delta) + k}{\nID \cdot N}\right)^\frac{1}{2} \right), 
    \end{align}
    where $C = \frac{\pi}{2} \cdot \left(\frac{3 \pi}{2} \right)^\frac{1}{2}$. 
\end{theorem}

\begin{proof}
    Let $\sS_0^{\textnormal{ID}},\sS_1^{\textnormal{ID}}$ denotes the hyperspheres for ID clients and $\sS_0^{\textnormal{OOD}},\sS_1^{\textnormal{OOD}}$ denotes the hyperspheres for OOD clients, by $\forall y, y' \in \{0, 1\}, \forall \vf_{1}, \vf_2 \in \sS_y^{\textnormal{ID}}, \vf_3 \in \sS_{y'}^{\textnormal{OOD}}$
    \begin{align*}
        \| \vf_1 - \vf_2 \|_2 < \| \vf_1 - \vf_3 \|_2. 
    \end{align*}
    Since we proved in Lemma \ref{lem:1nn} that {\algname} is equivalent to an 1-NN classifier, as long as there are ID samples in the merged memory, OOD samples will not be selected as the nearest. Therefore, the existence of OOD does not affect the behavior of the classifier. Equivalently, {\algname} selects no more than $k$ samples from in total $\nID \cdot N$ samples. We can apply Theorem \ref{thm:single_client} to get the final bound. 
\end{proof}

\newpage
\section{Additional Experiments}

\subsection{Detailed Experimental Setup}
\label{appendix:exp:setup}

\paragraph{Datasets} 
We summarize the statistics of datasets in Table \ref{tab:dataset_stats}. For domain adaptation benchmarks (VLCS, TerraIncognita), we use the implementation in DomainBed \cite{domainbed}. For corruption benchmarks, we use the code provided in \cite{corruption} to process both the original training and testing set of CIFAR-10 and CIFAR-100, thus the number of samples is 60,000 instead of 10,000. 

\begin{table*}[h!]
    \centering
    
    \small
        \begin{tabular}{lcccccccc}
            \toprule
            Dataset & \#Classes $c$ & \#Domains & \#Samples & \#Clients per Domain $m$ & \#Clients $n$\\
            \midrule
            VLCS            & 5     & 4     & 10,729    & 10    & 40 \\
            TerraIncognita  & 10    & 4     & 24,788    & 10    & 40 \\
            CIFAR-10-C      & 10    & 19    & 60,000    & 10    & 190 \\
            CIFAR-100-C     & 100   & 19    & 60,000    & 3     & 57 \\
            \bottomrule 
        \end{tabular}
    \caption{Dataset statistics} \label{tab:dataset_stats}
\end{table*}

\paragraph{Text templates and embeddings}

We use the 7 templates in \cite{tip-adaptor}: 
\begin{itemize}
    \item ``\texttt{itap of a \{class\}}''
    \item ``\texttt{a bad photo of the \{class\}}''
    \item ``\texttt{a origami \{class\}}''
    \item ``\texttt{a photo of the large \{class\}}''
    \item ``\texttt{a \{class\} in a video game}''  
    \item ``\texttt{art of the \{class\}}''
    \item ``\texttt{a photo of the small \{class\}}''
\end{itemize}
The text embedding for each class $y$ is computed by
\begin{align}
    \vt_y = \normalize\left( \sum_{\kappa = 1}^k \vt_{y, \kappa} \right), \text{where } \vt_{y, \kappa} = \normalize(\mathrm{text\_encoder(\{template_\kappa, classname_y\})})
\end{align}

\paragraph{Hyperparameter selection}
We adopt the same hyperparameter selection strategy as DMN \cite{dmn}, to both our proposed {\algname} and baseline TTA methods. To minimize the risk of overfitting and data leakage due to hyperparameter selection, we implement the following measures: For the corruption benchmark, we apply different corruptions during hyperparameter selection and testing, ensuring that the exact same images do not appear in both phases. For the domain adaptation benchmark, we use different partitions and random seeds to prevent overfitting to a specific data split, ensuring that the results remain representative and unbiased. Our final hyperparameters are: 
\begin{itemize}
    \item For VLCS, we use $\alpha = 0.3, \beta = 6.0, \gamma = 6.0, k_l = 15, k_e = 12$. 
    \item For TerraIncognita, we use $\alpha = 1.5, \beta = 35.0, \gamma = 10.0, k_l = 2, k_e = 20$. 
    \item For CIFAR-10-C, we use $\alpha = 1.0, \beta = 60.0, \gamma = 1.5, k_l = 12, k_e = 9$. 
    \item For CIFAR-100-C, we use $\alpha = 0.7, \beta = 60.0, \gamma = 1.5, k_l = 8, k_e = 5$. 
\end{itemize}


\newpage
\subsection{Additional Experiments}
\label{appendix:exp:additional}

\paragraph{Experiments on corruption benchmarks with RN50}

\begin{table*}[h!]
    \centering
    \small
    \resizebox{1.0\linewidth}{!}{
    \setlength{\tabcolsep}{1.0mm}{
        \begin{tabular}{lcccccccccccccccccccg}
            \toprule
            \multirow{4}{*}{Method} & \multicolumn{20}{c}{CIFAR-10-C} \\
            \cmidrule(lr){2-21}
            & \multicolumn{4}{c}{Noise} & \multicolumn{5}{c}{Blur} & \multicolumn{5}{c}{Weather} & \multicolumn{5}{c}{Digital} & \multirow{2.5}{*}{\cellcolor{White}Total} \\
            \cmidrule(lr){2-5} \cmidrule(lr){6-10} \cmidrule(lr){11-15} \cmidrule(lr){16-20}
            & Gauss. & Shot & Impulse & Speckle 
            & Defocus & Glass & Motion & Zoom & Gauss. 
            & Snow & Frost & Fog & Bright & Spatter
            & Contrast & Elastic & Pixel & JPEG & Saturate \\
            \midrule
            CLIP (RN50) \cite{clip} 
                & 19.21 & 22.15 & 16.83 & 23.99 & 37.42 & 26.51 & 32.54 & 39.49 & 33.25 & 50.47 & 51.43 & 46.36 & 59.80 & 60.40 & 27.08 & 34.99 & 30.95 & 42.75 & 56.88 & 37.50 \\
            VTE \cite{vte}            
                & 16.17 & 18.98 & 20.30 & 21.04 & 34.25 & 30.04 & 33.55 & 38.94 & 29.59 & 49.12 & 52.11 & 45.51 & 60.51 & 56.83 & 49.96 & 34.56 & 39.70 & 40.77 & 55.63 & 38.29 \\
            TPT \cite{tpt} 
                & 24.39 & 25.67 & 23.91 & 27.16 & 39.77 & 29.28 & 35.83 & 43.63 & 35.53 & 52.86 & 53.36 & 48.52 & 61.60 & 60.53 & 34.02 & 38.18 & 31.94 & 45.24 & 57.18 & 40.45 \\
            Zero \cite{zero}
                & 16.90 & 19.45 & 20.58 & 21.46 & 34.82 & 30.18 & 33.78 & 39.34 & 30.83 & 49.39 & 52.56 & 45.68 & 60.47 & 56.85 & 49.61 & 35.04 & 38.99 & 41.20 & 55.54 & 38.56 \\
            TDA (local) \cite{tda} 
                & 20.13 & 22.06 & 17.09 & 23.88 & 40.16 & 27.96 & 35.57 & 43.05 & 36.29 & 51.58 & 52.97 & 47.76 & 62.14 & 60.71 & 29.56 & 37.35 & 33.11 & 43.77 & 59.57 & 39.20 \\
            TDA (global) \cite{tda} 
                & 18.78 & 21.06 & 17.15 & 23.24 & 40.87 & 27.53 & 35.77 & 42.53 & 36.35 & 51.86 & 53.86 & 47.19 & 63.26 & 61.29 & 28.02 & 36.02 & 32.50 & 44.83 & 57.85 & 38.95 \\
            DMN-ZS (local) \cite{dmn}
                & 21.34 & 23.26 & 18.04 & 25.18 & 39.91 & 28.30 & 35.88 & 43.12 & 36.63 & 52.35 & 52.98 & 48.54 & 63.39 & 61.14 & 30.08 & 38.20 & 33.83 & 44.75 & 60.34 & 39.86 \\
            DMN-ZS (global) \cite{dmn}
                & 14.33 & 17.16 & 13.32 & 18.86 & 34.82 & 23.59 & 34.20 & 40.85 & 27.90 & 53.66 & 52.94 & 46.50 & 64.91 & 55.69 & 25.46 & 33.34 & 28.83 & 42.27 & 51.80 & 35.81 \\
            {\algname}
                & 21.49 & 23.38 & 18.80 & 25.41 & 40.14 & 29.00 & 36.67 & 43.69 & 37.37 & 52.10 & 52.34 & 49.22 & 63.90 & 59.42 & 30.88 & 38.99 & 34.17 & 43.95 & 60.28 & 40.06 \\
            \midrule \midrule
            \multirow{4}{*}{Method} & \multicolumn{20}{c}{CIFAR-100-C} \\
            \cmidrule(lr){2-21}
            & \multicolumn{4}{c}{Noise} & \multicolumn{5}{c}{Blur} & \multicolumn{5}{c}{Weather} & \multicolumn{5}{c}{Digital} & \multirow{2.5}{*}{\cellcolor{White}Total} \\
            \cmidrule(lr){2-5} \cmidrule(lr){6-10} \cmidrule(lr){11-15} \cmidrule(lr){16-20}
            & Gauss. & Shot & Impulse & Speckle 
            & Defocus & Glass & Motion & Zoom & Gauss. 
            & Snow & Frost & Fog & Bright & Spatter
            & Contrast & Elastic & Pixel & JPEG & Saturate \\
            \midrule
            CLIP (RN50) \cite{clip} 
                & 7.69  & 8.25  & 3.46  & 9.08 & 15.52  & 9.51 & 16.31 & 19.10 & 13.52 & 23.58 & 24.11 & 20.41 & 29.01 & 27.46 & 11.78 & 15.30 & 13.49 & 17.29 & 23.78 & 16.24 \\
            VTE \cite{vte}            
                & 6.87  & 7.65  & 3.59  & 7.98 & 12.52 & 10.97 & 14.14 & 15.04 & 10.83 & 19.63 & 20.52 & 18.13 & 23.19 & 23.70 & 20.71 & 13.89 & 13.80 & 14.41 & 19.08 & 14.56 \\
            TPT \cite{tpt}
                & 5.68  & 6.62  & 2.31  & 7.21 & 13.22 & 10.75 & 14.79 & 16.27 & 11.03 & 22.45 & 22.73 & 20.18 & 26.45 & 26.76 & 11.65 & 15.19 & 11.87 & 15.84 & 23.01 & 14.95 \\
            Zero \cite{zero}
                & 6.70  & 7.66  & 3.54  & 7.83 & 12.95 & 10.79 & 14.39 & 15.68 & 11.08 & 19.94 & 20.76 & 17.90 & 23.84 & 24.08 & 20.77 & 14.00 & 13.78 & 14.39 & 19.37 & 14.71 \\
            TDA (local) \cite{tda} 
                & 7.25  & 8.05  & 3.36  & 9.08 & 15.72  & 9.79 & 16.67 & 19.20 & 13.35 & 23.96 & 24.34 & 20.92 & 29.44 & 27.75 & 12.10 & 15.38 & 13.98 & 17.77 & 24.22 & 16.44 \\
            TDA (global) \cite{tda} 
                & 8.01  & 8.43  & 3.40  & 9.54 & 15.88  & 9.44 & 17.18 & 19.80 & 14.07 & 24.33 & 24.91 & 21.14 & 29.93 & 27.75 & 11.70 & 15.18 & 13.50 & 17.82 & 24.31 & 16.65 \\
            DMN-ZS (local) \cite{dmn} 
                & 7.61  & 8.30  & 3.63  & 9.27 & 15.82 & 10.01 & 16.54 & 19.52 & 13.71 & 24.17 & 24.61 & 21.11 & 29.77 & 27.89 & 11.61 & 15.76 & 13.90 & 17.74 & 24.50 & 16.60 \\
            DMN-ZS (global) \cite{dmn} 
                & 4.54  & 4.83  & 1.48  & 4.97 & 14.15  & 9.18 & 16.70 & 16.81 & 12.04 & 23.34 & 24.43 & 20.61 & 30.88 & 23.44  & 8.14 & 14.67  & 9.52 & 16.44 & 22.55 & 14.67 \\
            {\algname}
                & 7.72  & 8.44  & 3.71  & 9.44 & 16.13 & 10.32 & 17.01 & 19.68 & 14.08 & 24.46 & 24.84 & 21.30 & 30.03 & 27.95 & 11.96 & 15.91 & 14.11 & 17.84 & 24.84 & 16.83 \\
            \bottomrule
        \end{tabular}
    }
    }
    \caption{Accuracy (mean \% over five random seeds) on corruption benchmarks. }
    \label{tab:acc:corruption_rn}
\end{table*}

\paragraph{Additional hyperparameter sensitivity}

\begin{figure}[h!]
    \centering
    \includegraphics[width=0.6\linewidth]{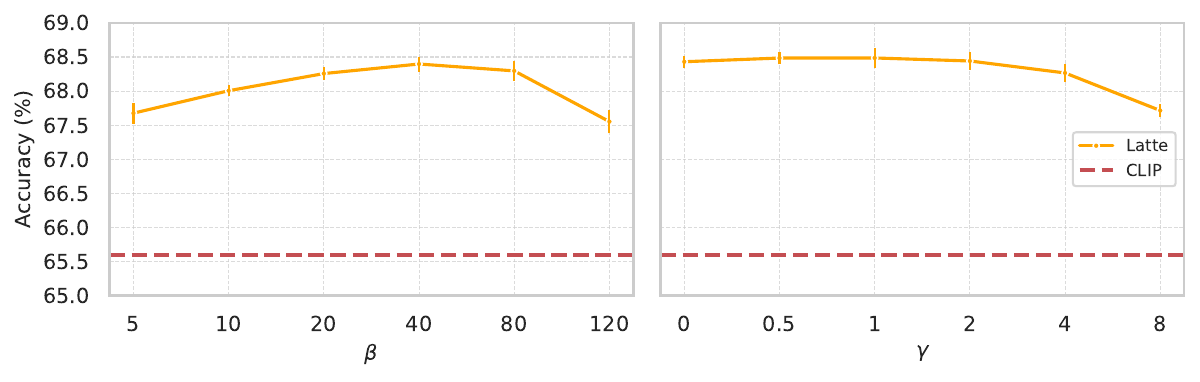}
    \caption{Hyperparameter sensitivity analysis on CIFAR-10-C. }
    \label{fig:hyperparameter_2}
\end{figure}

\paragraph{Additional experiments on levels of data decentralization and communication period}

\begin{figure}[h!]
    \centering
    \begin{minipage}[t]{0.45\linewidth}
        \centering
        \includegraphics[width=\linewidth]{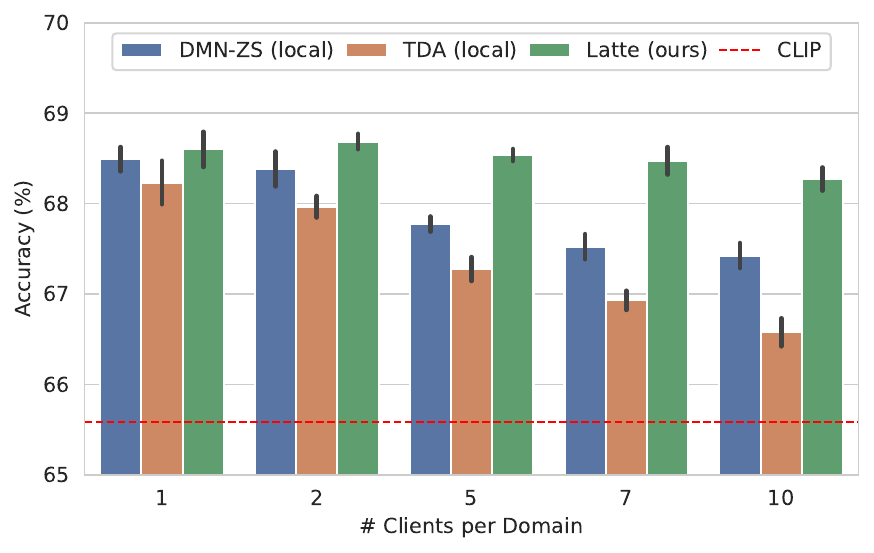}
        \caption{Comparison of memory-based TTA methods under different levels of data decentralization on CIFAR-10-C}
    \end{minipage}%
    \hspace{0.1\linewidth}
    \begin{minipage}[t]{0.4\linewidth}
        \centering
        \includegraphics[width=\linewidth]{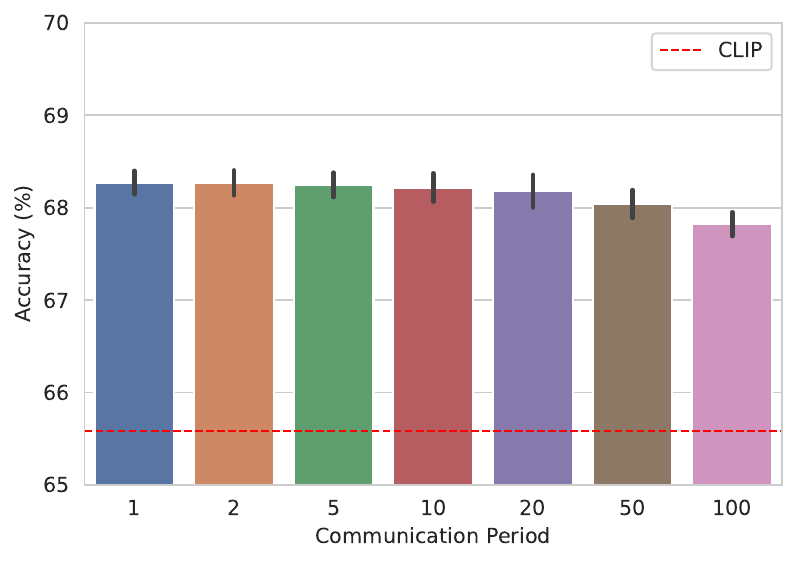}
        \caption{Performance of {\algname} with different communication period on CIFAR-10-C}
    \end{minipage}
    \vspace{-5ex}
\end{figure}


\newpage
\subsection{Standard Deviation Reporting}

Due to space limitations, we omitted standard deviation values in Tables \ref{tab:acc:domain}, \ref{tab:acc:corruption_vit}, and \ref{tab:acc:corruption_rn}. Here, we report the standard deviations, computed over five random permutations of the test sample stream. For methods such as CLIP, VTE, TPT, and Zero, which generate independent predictions for each sample, the results are unaffected by the ordering of the test stream. Therefore, we do not report their standard deviations, although their performance may still be influenced by randomness from augmentation.

\begin{table*}[h!]
    \centering
    \scriptsize
    \resizebox{1.0\linewidth}{!}{
    \setlength{\tabcolsep}{1.5mm}{
        \begin{tabular}{lcccccccccc}
            \toprule
            \multirow{2.5}{*}{Method} & \multicolumn{5}{c}{VLCS} & \multicolumn{5}{c}{TerraIncognita}  \\
            \cmidrule(lr){2-6} \cmidrule(lr){7-11} 
            & C & L & S & V & Total & L100 & L38 & L43 & L46 & Total \\
            \midrule
            CLIP (ViT-B/16) \cite{clip} 
                & 99.86 & 70.11 & 76.66 & 85.34 & 80.83
                & 41.96 & 28.30 & 35.82 & 26.82 & 31.84 \\
            TDA (local) \cite{tda} 
                & \meansd{99.94}{0.06} & \meansd{71.05}{0.28} & \meansd{77.82}{0.15} & \meansd{85.39}{0.19} & \meansd{81.44}{0.14} 
                & \meansd{40.46}{0.55} & \meansd{32.72}{0.50} & \meansd{36.23}{0.33} & \meansd{30.41}{0.40} & \meansd{34.24}{0.30} \\
            TDA (global) \cite{tda} 
                & \meansd{99.94}{0.03} & \meansd{67.53}{0.32} & \meansd{76.95}{0.05} & \meansd{85.34}{0.04} & \meansd{80.29}{0.08} 
                & \meansd{41.78}{1.46} & \meansd{39.58}{0.48} & \meansd{35.37}{1.04} & \meansd{26.87}{0.25} & \meansd{36.19}{0.47} \\
            DMN-ZS (local) \cite{dmn}
                & \meansd{99.90}{0.04} & \meansd{70.50}{0.30} & \meansd{77.58}{0.20} & \meansd{85.05}{0.16} & \meansd{81.12}{0.08} 
                & \meansd{37.79}{0.52} & \meansd{31.37}{1.67} & \meansd{40.70}{0.77} & \meansd{29.34}{0.72} & \meansd{33.65}{0.81} \\
            DMN-ZS (global) \cite{dmn}
                & \meansd{99.93}{0.00} & \meansd{68.03}{0.02} & \meansd{77.75}{0.03} & \meansd{85.00}{0.07} & \meansd{80.55}{0.03} 
                & \meansd{38.33}{0.33} & \meansd{44.39}{0.16} & \meansd{36.75}{2.39} & \meansd{26.91}{0.21} & \meansd{37.64}{0.42} \\
            {\algname}
                & \meansd{99.96}{0.04} & \meansd{73.00}{0.68} & \meansd{80.18}{0.40} & \meansd{85.13}{0.10} & \meansd{82.57}{0.12} 
                & \meansd{39.54}{2.39} & \meansd{51.54}{0.59} & \meansd{33.43}{1.85} & \meansd{30.12}{1.55} & \meansd{40.95}{1.06} \\
            \midrule
            \midrule

            CLIP (RN50) \cite{clip} 
                & 99.36 & 68.60 & 73.52 & 84.92 & 79.30 
                & 14.70 & 20.04 & 34.43 & 27.82 & 23.24 \\
            TDA (local) \cite{tda} 
                & \meansd{99.56}{0.09} & \meansd{69.32}{0.47} & \meansd{74.59}{0.17} & \meansd{84.77}{0.08} & \meansd{79.78}{0.15} 
                & \meansd{29.68}{0.49} & \meansd{40.39}{0.74} & \meansd{38.80}{0.30} & \meansd{35.14}{0.61} & \meansd{36.73}{0.18} \\
            TDA (global) \cite{tda} 
                & \meansd{99.59}{0.03} & \meansd{67.45}{0.12} & \meansd{74.16}{0.04} & \meansd{84.56}{0.04} & \meansd{79.12}{0.04} 
                & \meansd{29.28}{3.39} & \meansd{43.16}{1.10} & \meansd{39.58}{0.15} & \meansd{34.80}{0.17} & \meansd{37.78}{1.00} \\
            DMN-ZS (local) \cite{dmn}
                & \meansd{99.48}{0.06} & \meansd{69.05}{0.27} & \meansd{73.92}{0.21} & \meansd{84.66}{0.11} & \meansd{79.47}{0.12} 
                & \meansd{25.81}{1.34} & \meansd{32.52}{2.32} & \meansd{37.45}{0.69} & \meansd{32.07}{0.82} & \meansd{31.89}{1.02} \\
            DMN-ZS (global) \cite{dmn}
                & \meansd{99.56}{0.09} & \meansd{65.89}{0.08} & \meansd{74.77}{0.08} & \meansd{83.96}{0.09} & \meansd{78.73}{0.05} 
                & \meansd{25.68}{2.61} & \meansd{44.35}{1.27} & \meansd{36.71}{0.28} & \meansd{29.33}{0.10} & \meansd{35.73}{1.02} \\
            {\algname} 
                & \meansd{99.66}{0.03} & \meansd{72.07}{0.41} & \meansd{77.19}{0.53} & \meansd{83.12}{0.29} & \meansd{80.75}{0.22} 
                & \meansd{47.51}{0.83} & \meansd{45.18}{0.78} & \meansd{37.38}{0.43} & \meansd{33.40}{0.42} & \meansd{41.47}{0.51} \\
            
            \bottomrule 
        \end{tabular}
        }}
    \caption{Accuracy (\meansd{mean}{s.d.} \% over five random seeds) on domain adaptation benchmarks.}
    \label{tab:acc:domain_sd}
\end{table*}


\begin{table*}[h!]
    \centering
    \small
    \resizebox{1.0\linewidth}{!}{
    \setlength{\tabcolsep}{1.0mm}{
        \begin{tabular}{lcccccccccccccccccccc}
            \toprule
            & \multicolumn{20}{c}{CIFAR-10-C} \\
            \cmidrule(lr){2-21}
            Method & \multicolumn{4}{c}{Noise} & \multicolumn{5}{c}{Blur} & \multicolumn{5}{c}{Weather} & \multicolumn{5}{c}{Digital} & \multirow{2.5}{*}{Total} \\
            \cmidrule(lr){2-5} \cmidrule(lr){6-10} \cmidrule(lr){11-15} \cmidrule(lr){16-20}
            & Gauss. & Shot & Impulse & Speckle 
            & Defocus & Glass & Motion & Zoom & Gauss. 
            & Snow & Frost & Fog & Bright & Spatter
            & Contrast & Elastic & Pixel & JPEG & Saturate \\
            \midrule
            CLIP (ViT-B/16) \cite{clip} 
                & 40.89 & 45.36 & 58.14 & 48.66 & 73.20 & 44.20 & 70.21 & 75.33 & 71.30 & 76.34 & 79.88 & 72.12 & 85.93 & 84.38 & 65.56 & 55.93 & 51.05 & 61.59 & 85.93 & 65.58 \\
            TDA (local) \cite{tda} 
                & \meansd{40.72}{0.31} & \meansd{45.85}{0.48} & \meansd{59.85}{1.01} & \meansd{48.83}{0.60} & \meansd{74.23}{0.58} & \meansd{45.54}{1.28} & \meansd{71.38}{0.78} & \meansd{76.61}{0.41} & \meansd{72.28}{0.58} & \meansd{77.19}{0.51} & \meansd{80.55}{0.56} & \meansd{73.11}{0.72} & \meansd{86.39}{0.29} & \meansd{84.94}{0.66} & \meansd{67.68}{1.23} & \meansd{57.97}{0.82} & \meansd{52.84}{0.53} & \meansd{62.52}{1.12} & \meansd{86.43}{0.80} & \meansd{66.58}{0.15} \\
            TDA (global) \cite{tda} 
                & \meansd{39.46}{1.35} & \meansd{44.81}{1.09} & \meansd{56.56}{1.07} & \meansd{47.93}{1.02} & \meansd{73.05}{0.40} & \meansd{44.80}{1.34} & \meansd{70.67}{0.70} & \meansd{75.79}{0.60} & \meansd{70.94}{0.78} & \meansd{76.90}{0.66} & \meansd{80.53}{0.65} & \meansd{72.65}{0.54} & \meansd{86.31}{0.62} & \meansd{85.00}{0.59} & \meansd{65.61}{0.65} & \meansd{56.35}{1.10} & \meansd{50.84}{0.66} & \meansd{61.71}{0.66} & \meansd{86.01}{0.98} & \meansd{65.58}{0.29} \\
            DMN-ZS (local) \cite{dmn}
                & \meansd{42.79}{0.61} & \meansd{47.27}{0.50} & \meansd{60.46}{0.50} & \meansd{50.46}{0.67} & \meansd{75.09}{0.50} & \meansd{46.00}{0.82} & \meansd{72.23}{0.94} & \meansd{77.39}{0.45} & \meansd{73.58}{0.91} & \meansd{77.74}{0.49} & \meansd{80.60}{0.36} & \meansd{73.54}{0.51} & \meansd{87.13}{0.40} & \meansd{85.74}{0.52} & \meansd{68.21}{0.94} & \meansd{58.54}{0.85} & \meansd{54.12}{0.81} & \meansd{63.29}{1.02} & \meansd{86.87}{0.55} & \meansd{67.42}{0.14} \\
            DMN-ZS (global) \cite{dmn}
                & \meansd{35.60}{1.27} & \meansd{40.56}{1.95} & \meansd{52.05}{1.40} & \meansd{42.85}{1.72} & \meansd{72.10}{0.36} & \meansd{44.20}{1.33} & \meansd{68.93}{0.53} & \meansd{74.92}{0.67} & \meansd{69.25}{0.73} & \meansd{76.22}{0.24} & \meansd{80.01}{0.77} & \meansd{72.24}{0.60} & \meansd{86.36}{0.47} & \meansd{84.62}{0.80} & \meansd{63.02}{1.31} & \meansd{57.42}{1.17} & \meansd{50.56}{1.96} & \meansd{58.44}{1.49} & \meansd{84.80}{0.70} & \meansd{63.90}{0.40} \\
            {\algname}
                & \meansd{43.72}{0.68} & \meansd{48.46}{0.68} & \meansd{61.67}{0.70} & \meansd{51.21}{0.40} & \meansd{76.19}{0.59} & \meansd{46.86}{1.02} & \meansd{73.10}{1.01} & \meansd{78.75}{0.72} & \meansd{74.69}{0.81} & \meansd{78.37}{0.50} & \meansd{81.22}{0.63} & \meansd{73.67}{0.71} & \meansd{87.38}{0.57} & \meansd{86.51}{0.45} & \meansd{68.43}{1.21} & \meansd{60.22}{0.97} & \meansd{55.76}{0.88} & \meansd{63.67}{0.78} & \meansd{87.29}{0.59} & \meansd{68.27}{0.13} \\
            \midrule \midrule
            CLIP (RN50) \cite{clip} 
                & 19.21 & 22.15 & 16.83 & 23.99 & 37.42 & 26.51 & 32.54 & 39.49 & 33.25 & 50.47 & 51.43 & 46.36 & 59.80 & 60.40 & 27.08 & 34.99 & 30.95 & 42.75 & 56.88 & 37.50 \\
            TDA (local) \cite{tda} 
                & \meansd{20.13}{1.02} & \meansd{22.06}{0.70} & \meansd{17.09}{0.43} & \meansd{23.88}{0.68} & \meansd{40.16}{0.29} & \meansd{27.96}{0.99} & \meansd{35.57}{0.71} & \meansd{43.05}{0.41} & \meansd{36.29}{0.68} & \meansd{51.58}{0.61} & \meansd{52.97}{0.83} & \meansd{47.76}{0.40} & \meansd{62.14}{1.01} & \meansd{60.71}{0.49} & \meansd{29.56}{1.01} & \meansd{37.35}{1.02} & \meansd{33.11}{1.23} & \meansd{43.77}{0.85} & \meansd{59.57}{0.43} & \meansd{39.20}{0.23} \\
            TDA (global) \cite{tda} 
                & \meansd{18.78}{1.06} & \meansd{21.06}{0.70} & \meansd{17.15}{0.76} & \meansd{23.24}{0.94} & \meansd{40.87}{2.90} & \meansd{27.53}{1.74} & \meansd{35.77}{2.68} & \meansd{42.53}{1.58} & \meansd{36.35}{2.85} & \meansd{51.86}{1.29} & \meansd{53.86}{1.69} & \meansd{47.19}{1.24} & \meansd{63.26}{1.25} & \meansd{61.29}{0.95} & \meansd{28.02}{1.30} & \meansd{36.02}{1.16} & \meansd{32.50}{1.38} & \meansd{44.83}{1.38} & \meansd{57.85}{1.07} & \meansd{38.95}{1.12} \\
            DMN-ZS (local) \cite{dmn}
                & \meansd{21.34}{0.63} & \meansd{23.26}{0.70} & \meansd{18.04}{0.57} & \meansd{25.18}{0.57} & \meansd{39.91}{0.39} & \meansd{28.30}{1.08} & \meansd{35.88}{0.67} & \meansd{43.12}{0.57} & \meansd{36.63}{0.47} & \meansd{52.35}{0.75} & \meansd{52.98}{0.89} & \meansd{48.54}{0.34} & \meansd{63.39}{0.79} & \meansd{61.14}{0.42} & \meansd{30.08}{0.81} & \meansd{38.20}{0.91} & \meansd{33.83}{0.60} & \meansd{44.75}{1.21} & \meansd{60.34}{0.47} & \meansd{39.86}{0.21} \\
            DMN-ZS (global) \cite{dmn}
                & \meansd{14.33}{1.37} & \meansd{17.16}{2.57} & \meansd{13.32}{0.89} & \meansd{18.86}{2.14} & \meansd{34.82}{2.95} & \meansd{23.59}{1.56} & \meansd{34.20}{0.87} & \meansd{40.85}{3.58} & \meansd{27.90}{2.17} & \meansd{53.66}{1.06} & \meansd{52.94}{1.04} & \meansd{46.50}{1.33} & \meansd{64.91}{0.97} & \meansd{55.69}{1.18} & \meansd{25.46}{2.69} & \meansd{33.34}{1.59} & \meansd{28.83}{1.30} & \meansd{42.27}{1.63} & \meansd{51.80}{0.90} & \meansd{35.81}{0.69} \\
            {\algname}
                & \meansd{21.49}{0.63} & \meansd{23.38}{0.54} & \meansd{18.80}{0.80} & \meansd{25.41}{0.90} & \meansd{40.14}{0.47} & \meansd{29.00}{1.26} & \meansd{36.67}{1.08} & \meansd{43.69}{0.57} & \meansd{37.37}{0.58} & \meansd{52.10}{0.95} & \meansd{52.34}{1.04} & \meansd{49.22}{0.76} & \meansd{63.90}{0.65} & \meansd{59.42}{0.58} & \meansd{30.88}{0.83} & \meansd{38.99}{0.74} & \meansd{34.17}{0.91} & \meansd{43.95}{1.20} & \meansd{60.28}{0.56} & \meansd{40.06}{0.13} \\
            \midrule \midrule
            & \multicolumn{20}{c}{CIFAR-100-C} \\
            \cmidrule(lr){2-21}
            Method & \multicolumn{4}{c}{Noise} & \multicolumn{5}{c}{Blur} & \multicolumn{5}{c}{Weather} & \multicolumn{5}{c}{Digital} & \multirow{2.5}{*}{Total} \\
            \cmidrule(lr){2-5} \cmidrule(lr){6-10} \cmidrule(lr){11-15} \cmidrule(lr){16-20}
            & Gauss. & Shot & Impulse & Speckle 
            & Defocus & Glass & Motion & Zoom & Gauss. 
            & Snow & Frost & Fog & Bright & Spatter
            & Contrast & Elastic & Pixel & JPEG & Saturate \\
            \midrule
            CLIP (ViT-B/16) \cite{clip} 
                & 22.56 & 23.97 & 30.16 & 24.91 & 44.03 & 20.25 & 43.53 & 48.54 & 42.67 & 49.52 & 51.08 & 41.93 & 57.94 & 57.61 & 35.31 & 29.63 & 26.09 & 33.45 & 56.06 & 38.91 \\
            TDA (local) \cite{tda} 
                & \meansd{22.39}{0.69} & \meansd{24.09}{0.46} & \meansd{30.17}{0.44} & \meansd{24.83}{0.58} & \meansd{44.31}{0.68} & \meansd{20.27}{0.31} & \meansd{43.95}{0.73} & \meansd{48.94}{0.51} & \meansd{42.89}{0.53} & \meansd{49.54}{0.64} & \meansd{51.23}{0.35} & \meansd{41.98}{0.94} & \meansd{58.21}{0.41} & \meansd{58.06}{0.51} & \meansd{35.35}{0.83} & \meansd{29.73}{0.78} & \meansd{26.17}{0.80} & \meansd{33.51}{0.59} & \meansd{56.40}{0.54} & \meansd{39.05}{0.10} \\
            TDA (global) \cite{tda} 
                & \meansd{22.65}{0.60} & \meansd{24.33}{0.83} & \meansd{30.21}{0.41} & \meansd{24.98}{0.73} & \meansd{44.60}{0.56} & \meansd{19.77}{0.42} & \meansd{43.68}{0.75} & \meansd{48.81}{0.30} & \meansd{42.20}{0.57} & \meansd{49.36}{0.64} & \meansd{51.31}{0.41} & \meansd{41.90}{0.88} & \meansd{58.28}{0.39} & \meansd{57.88}{0.64} & \meansd{35.36}{0.75} & \meansd{29.61}{0.72} & \meansd{26.21}{0.83} & \meansd{33.44}{0.37} & \meansd{56.28}{0.43} & \meansd{38.99}{0.13} \\
            DMN-ZS (local) \cite{dmn}
                & \meansd{23.00}{0.73} & \meansd{24.36}{0.69} & \meansd{30.88}{0.67} & \meansd{25.15}{0.59} & \meansd{44.34}{0.47} & \meansd{20.51}{0.37} & \meansd{43.68}{0.78} & \meansd{48.82}{0.57} & \meansd{43.01}{0.77} & \meansd{49.63}{0.65} & \meansd{51.14}{0.48} & \meansd{42.08}{1.02} & \meansd{58.28}{0.39} & \meansd{57.87}{0.53} & \meansd{35.38}{0.77} & \meansd{29.89}{0.83} & \meansd{26.37}{0.58} & \meansd{33.53}{0.60} & \meansd{56.31}{0.67} & \meansd{39.17}{0.16} \\
            DMN-ZS (global) \cite{dmn} 
                & \meansd{18.85}{1.94} & \meansd{20.38}{0.80} & \meansd{25.32}{2.42} & \meansd{21.42}{0.99} & \meansd{42.39}{0.32} & \meansd{16.17}{0.73} & \meansd{42.96}{0.72} & \meansd{47.01}{0.43} & \meansd{39.83}{1.33} & \meansd{47.51}{0.99} & \meansd{49.76}{1.04} & \meansd{40.15}{1.24} & \meansd{59.07}{0.25} & \meansd{57.09}{0.42} & \meansd{30.64}{0.26} & \meansd{26.78}{1.04} & \meansd{23.15}{0.19} & \meansd{31.29}{0.88} & \meansd{56.35}{0.90} & \meansd{36.64}{0.50} \\
            {\algname} 
                & \meansd{23.34}{0.65} & \meansd{24.74}{0.51} & \meansd{31.36}{0.59} & \meansd{25.26}{0.51} & \meansd{44.97}{0.54} & \meansd{21.08}{0.45} & \meansd{44.03}{0.74} & \meansd{49.58}{0.74} & \meansd{43.32}{0.48} & \meansd{49.66}{0.78} & \meansd{51.45}{0.31} & \meansd{41.99}{1.00} & \meansd{58.64}{0.41} & \meansd{58.51}{0.62} & \meansd{35.20}{0.65} & \meansd{30.16}{1.03} & \meansd{26.78}{0.77} & \meansd{33.80}{0.49} & \meansd{56.87}{0.69} & \meansd{39.51}{0.14} \\
            \midrule \midrule
            CLIP (RN50) \cite{clip} 
                & 7.69  & 8.25  & 3.46  & 9.08 & 15.52  & 9.51 & 16.31 & 19.10 & 13.52 & 23.58 & 24.11 & 20.41 & 29.01 & 27.46 & 11.78 & 15.30 & 13.49 & 17.29 & 23.78 & 16.24 \\
            TDA (local) \cite{tda} 
                & \meansd{7.25}{0.40} & \meansd{8.05}{0.28} & \meansd{3.36}{0.30} & \meansd{9.08}{0.25} & \meansd{15.72}{0.25} & \meansd{9.79}{0.50} & \meansd{16.67}{0.51} & \meansd{19.20}{0.37} & \meansd{13.35}{0.50} & \meansd{23.96}{0.23} & \meansd{24.34}{0.65} & \meansd{20.92}{0.58} & \meansd{29.44}{0.76} & \meansd{27.75}{0.77} & \meansd{12.10}{0.51} & \meansd{15.38}{0.64} & \meansd{13.98}{0.42} & \meansd{17.77}{0.61} & \meansd{24.22}{0.81} & \meansd{16.44}{0.16} \\
            TDA (global) \cite{tda} 
                & \meansd{8.01}{0.60} & \meansd{8.43}{0.32} & \meansd{3.40}{0.40} & \meansd{9.54}{0.20} & \meansd{15.88}{0.35} & \meansd{9.44}{0.56} & \meansd{17.18}{0.30} & \meansd{19.80}{0.80} & \meansd{14.07}{0.54} & \meansd{24.33}{0.51} & \meansd{24.91}{0.75} & \meansd{21.14}{0.51} & \meansd{29.93}{0.66} & \meansd{27.75}{0.72} & \meansd{11.70}{0.50} & \meansd{15.18}{0.40} & \meansd{13.50}{0.28} & \meansd{17.82}{0.47} & \meansd{24.31}{1.16} & \meansd{16.65}{0.23} \\
            DMN-ZS (local) \cite{dmn}
                & \meansd{7.61}{0.49} & \meansd{8.30}{0.37} & \meansd{3.63}{0.39} & \meansd{9.27}{0.26} & \meansd{15.82}{0.14} & \meansd{10.01}{0.61} & \meansd{16.54}{0.53} & \meansd{19.52}{0.70} & \meansd{13.71}{0.64} & \meansd{24.17}{0.42} & \meansd{24.61}{0.72} & \meansd{21.11}{0.57} & \meansd{29.77}{0.82} & \meansd{27.89}{0.97} & \meansd{11.61}{0.45} & \meansd{15.76}{0.57} & \meansd{13.90}{0.33} & \meansd{17.74}{0.53} & \meansd{24.50}{0.61} & \meansd{16.60}{0.13} \\
            DMN-ZS (global) \cite{dmn}
                & \meansd{4.54}{0.36} & \meansd{4.83}{0.89} & \meansd{1.48}{0.17} & \meansd{4.97}{0.91} & \meansd{14.15}{0.75} & \meansd{9.18}{0.68} & \meansd{16.70}{0.42} & \meansd{16.81}{1.02} & \meansd{12.04}{0.79} & \meansd{23.34}{0.65} & \meansd{24.43}{0.70} & \meansd{20.61}{0.81} & \meansd{30.88}{0.47} & \meansd{23.44}{1.29} & \meansd{8.14}{0.59} & \meansd{14.67}{0.34} & \meansd{9.52}{0.45} & \meansd{16.44}{1.12} & \meansd{22.55}{0.81} & \meansd{14.67}{0.20} \\
            {\algname} 
                & \meansd{7.72}{0.38} & \meansd{8.44}{0.33} & \meansd{3.71}{0.36} & \meansd{9.44}{0.46} & \meansd{16.13}{0.24} & \meansd{10.32}{0.65} & \meansd{17.01}{0.52} & \meansd{19.68}{0.67} & \meansd{14.08}{0.56} & \meansd{24.46}{0.25} & \meansd{24.84}{0.62} & \meansd{21.30}{0.54} & \meansd{30.03}{0.87} & \meansd{27.95}{0.94} & \meansd{11.96}{0.40} & \meansd{15.91}{0.67} & \meansd{14.11}{0.37} & \meansd{17.84}{0.41} & \meansd{24.84}{0.59} & \meansd{16.83}{0.17} \\
            \bottomrule 
        \end{tabular}
    }
    }
    \caption{Accuracy (\meansd{mean}{s.d.} \% over five random seeds) on corruption benchmarks. }
    \label{tab:acc:corruption_sd}
\end{table*}


\end{document}